\documentclass[sigconf]{acmart}
\usepackage[utf8]{inputenc} 
\usepackage[T1]{fontenc}    
\usepackage{url}            
\usepackage{booktabs}       
\usepackage{amsfonts}       
\usepackage{nicefrac}       
\usepackage{microtype}      
\usepackage{amsthm}
\usepackage{amsmath}
\usepackage{graphicx}

\usepackage{amssymb}
\usepackage{bbm}
\usepackage{diagbox}
\usepackage{colortbl}
\usepackage{multirow}
\usepackage{pifont} 
\usepackage{bm}
\usepackage{subcaption}
\usepackage{color,xcolor}
\usepackage[font=small,labelfont=bf]{caption}  
\usepackage{makecell}
\usepackage{tabulary}
\definecolor{demphcolor}{RGB}{144,144,144}

\definecolor{mygray}{gray}{0.4}
\usepackage{colortbl}
\usepackage{enumitem}
\usepackage{multirow}
\usepackage{threeparttable}
\usepackage[export]{adjustbox}

\usepackage{makecell}
\usepackage{xcolor} 
\definecolor{snowblue}{rgb}{0.65, 0.80, 0.98} 

\usepackage{newfloat}
\usepackage{listings}
\AtBeginDocument{%
  }

\setcopyright{acmlicensed}
\copyrightyear{2024}
\acmYear{2024}
\acmConference[KDD '24]{Proceedings of the 30th ACM SIGKDD Conference on Knowledge Discovery and Data Mining}{August 25--29, 2024}{Barcelona, Spain}
\acmBooktitle{Proceedings of the 30th ACM SIGKDD Conference on Knowledge Discovery and Data Mining (KDD '24), August 25--29, 2024, Barcelona, Spain}
\acmDOI{10.1145/3637528.3671791}
\acmISBN{979-8-4007-0490-1/24/08}
\renewcommand{\shortauthors}{Kun Wang et al.}

\usepackage{twemojis}
\usepackage[linesnumbered,ruled,vlined]{algorithm2e} 
\SetKwInOut{Input}{Input}\SetKwInOut{Output}{Output}

\SetCommentSty{mycommfont}

\begin{document}

\title{The Heterophilic Snowflake Hypothesis: \\Training and Empowering GNNs for Heterophilic Graphs}


\author{Kun Wang\ding{94}}
\authornote{Kun Wang and Guibin Zhang are co-first authors. $\dagger$ denotes corresponding authors.}
\email{wk520529@mail.ustc.edu.cn}
\affiliation{\institution{University of Science and Technology of China (USTC)} \city{Hefei}\country{China} }

\author{Guibin Zhang\ding{94}}
\authornotemark[1]
\email{bin2003@tongji.edu.cn}
\affiliation{\institution{Tongji University}\city{Shanghai}\country{China}}

\author{Xinnan Zhang}
\email{zhan9359@umn.edu}
\affiliation{\institution{University of Minnesota}\city{Twin Cities}\country{USA}}

\author{Junfeng Fang}
\email{fjf@mail.ustc.edu.cn}
\affiliation{\institution{University of Science and Technology of China (USTC)}\city{Hefei}\country{China}}

\author{Xun Wu}
\email{wuxun21@mails.tsinghua.edu.cn}
\affiliation{\institution{Tsinghua University}\city{Beijing}\country{China}}

\author{Guohao Li}
\email{guohao.li@kaust.edu.sa}
\affiliation{\institution{Oxford University}\city{Oxford}\country{UK}}

\author{Shirui Pan}
\email{s.pan@griffith.edu.au}
\affiliation{\institution{Griffith University}\city{Queensland}\country{Austrilia}}

\author{Wei Huang$\dagger$}
\email{wei.huang.vr@riken.jp}
\affiliation{\institution{RIKEN AIP}\city{Tokyo}\country{Japan}}

\author{Yuxuan Liang$\dagger$}
\email{yuxliang@outlook.com}
\affiliation{\institution{The Hong Kong University of Science and Technology (Guangzhou)}\city{Guangzhou}\country{China}}

\renewcommand{\shortauthors}{Kun Wang et al.}

\begin{abstract}
    Graph Neural Networks (GNNs) have become pivotal tools for a range of graph-based learning tasks. Notably, most current GNN architectures operate under the assumption of homophily, whether explicitly or implicitly. While this underlying assumption is frequently adopted, it is not universally applicable, which can result in potential shortcomings in learning effectiveness. In this paper, \textbf{for the first time}, we transfer the prevailing concept of ``one node one receptive field" to the heterophilic graph. By constructing a proxy label predictor, we enable each node to possess a latent prediction distribution, which assists connected nodes in determining whether they should aggregate their associated neighbors. Ultimately, every node can have its own unique aggregation hop and pattern, much like each snowflake is unique and possesses its own characteristics. Based on observations, we innovatively introduce the Heterophily Snowflake Hypothesis and provide an effective solution to guide and facilitate research on heterophilic graphs and beyond.
    We conduct comprehensive experiments including (1) main results on 10 graphs with varying heterophily ratios across 10 backbones; (2) scalability on various deep GNN backbones (SGC, JKNet, etc.) across various large number of layers (2,4,6,8,16,32 layers); (3) comparison with conventional snowflake hypothesis; (4) efficiency comparison with existing graph pruning algorithms. Our observations show that our framework acts as a versatile operator for diverse tasks. It can be integrated into various GNN frameworks, boosting performance in-depth and offering an explainable approach to choosing the optimal network depth. The source code is available at \url{https://github.com/bingreeky/HeteroSnoH}.
    
\end{abstract}



\maketitle

\vspace{-0.6em}
\section{Introduction}\label{sec:intro}

Graph Neural Networks (GNNs)~\cite{kipf2016semi, hamilton2017inductive, wu2020comprehensive} have become the \textit{de facto} standard for various graph representation learning tasks, such as node classification \cite{velickovic2017graph, abu2020n}, link prediction \cite{zhang2018link, zhang2019inductive}, and graph classification \cite{ying2018hierarchical, gao2019graph}. The superior capabilities of GNNs can be attributed to their message passing paradigm \cite{wu2020comprehensive}. Through iterative information aggregation and updating, the central node captures rich information by interacting with its neighboring nodes based on the connected graph structure \cite{wu2020comprehensive, vignac2020building, zhou2020graph}.

\begin{figure}[!t]
  \centering
  \includegraphics[width=0.95\linewidth]{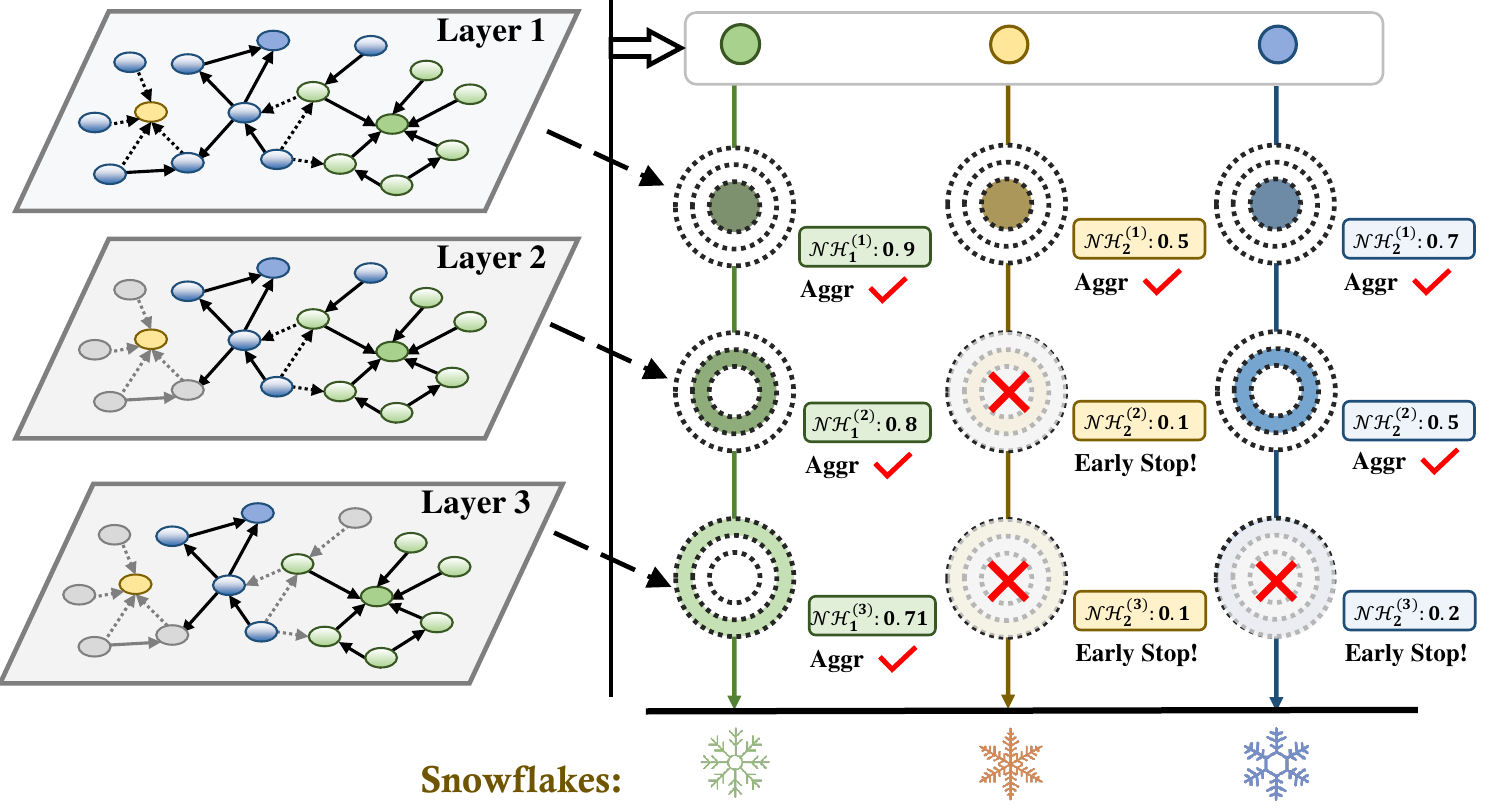}
  \vspace{-0.8em}
  \caption{The algorithm workflow of Heterophilic Snowflake Hypothesis (Hetero-S) and Heterophily-aware Early Stopping (HES). }
   \label{fig:intro}
\vspace{-0.8em}
\end{figure}

Among the various landscape of GNN architectures and designs, the homophily assumption \cite{zheng2022graph, zhu2021graph, li2022finding} serves as a foundational pillar, suggesting that edges predominantly link nodes with identical labels and analogous node features. Despite its appealing success, the performance of current GNNs has dropped sharply as the homophily of the graph decreases. Specifically, within heterophilic graphs, a discrepancy is often observed between the labels of neighboring nodes and the central node, a phenomenon referred to as the local structure discrepancy issue \cite{chen2020measuring, zheng2022graph, zhu2020beyond, luan2022revisiting}. We ascribe the observed performance degradation to the uniform message passing framework, in which that the central node initially aggregates messages from its local neighboring nodes, subsequently updating the ego node (see Figure \ref{fig:intro} left hand). 

However, non-euclidean data frequently display heterophily, which can be observed across diverse domains. For instance, when users engage with content on Netflix, the people with diverse preferences might be subjected to similar recommendation algorithms, owing to their interaction with identical video content. The potential of heterophilic graphs is vast, holding promise in both academic and industrial spheres, such as social networks \cite{fan2019graph}, transportation systems \cite{wang2022a2djp, zhou2022greto}, and recommendation platforms \cite{wu2022graph, wu2019session}. In a heterophilic context, this mechanism introduces two primary and challenging limitations:

\begin{itemize}[leftmargin=*]
    \item In graph topology, local neighbors refer to nearby nodes, often overlooking distant yet informative nodes. In heterophilic, nodes sharing structural and semantic characteristics can be more distantly positioned \cite{lim2021large, lim2021new}.

    \item A consistent aggregation and update method often neglects variations in information from alike/unalike neighbors.  In heterophilic graphs, achieving discerning node representations necessitates customized message passing to capture distinctive patterns.
    
\end{itemize}

\noindent Given the above emerging challenges, there has been a shifting focus towards exploring heterophily in GNNs. This research area includes a wide range from delving into heterophilic graph sampling to the evolution of intricate algorithms. The growing interest in heterophilic graph learning can be attributed to its vast applicability. 

From a macro perspective, existing heterophilic GNNs can be broadly classified into two categories, \textit{i.e.,} \textbf{non-local neighbor extension} and \textbf{GNN architecture refinement} \cite{zheng2022graph}. The concept of non-local neighbor extension in heterophilic GNNs involves broadening the receptive field beyond local neighbors. This is achieved through strategies like high-order neighbor information mixing \cite{abu2019mixhop, zhu2020beyond, jin2021universal, wang2021tree} and potential neighbor discovery \cite{fu2022p, liu2021non, yang2022graph, pei2020geom}, enhancing representation by capturing distant but relevant node features. With the second class, GNN architecture refinement focuses on enhancing the expressive capability of GNNs for heterophilic graphs by optimizing AGGREGATE and UPDATE function \cite{wu2020comprehensive}.  Through strategies such as adaptive message aggregation \cite{velivckovic2017graph, gasteiger2018predict}, ego-neighbor separation 
\cite{zhu2020beyond, suresh2021breaking}, and layer-wise operation \cite{xu2018representation, chen2020simple, chien2020adaptive}, the refinement aims to produce distinguishable and discriminative node representations.

Recently, a novel paradigm, \textit{the Snowflake Hypothesis} (SnoH)~\cite{wang2023snowflake}, rooted in the concept of ``one node, one receptive field'' has gained significant attention for its efficacy in addressing the over-smoothing \cite{rong2019dropedge} and over-fitting \cite{li2018deeper, chen2020measuring} issues in GNNs. This hypothesis draws inspiration from the intricate and distinctive patterns exhibited by individual snowflakes, assuming that each node can possess its unique receptive field. It posits that for an $L$-layer GNN, every node in the graph harbors an optimal receptive field width denoted as $k$ ($1 \leq k\leq L$). During the message passing process from $1$ to $L$ hops, each node merely aggregates information from the preceding $k$ hops, after which they cease to aggregate information from the neighborhood (referred to as ``\textit{node early stopping}''). 




Intuitively, the snowflake hypothesis demonstrates even greater vitality and significance in the context of heterophilic graphs: (1) \textbf{One concept benefits all.} The idea behind the snowflake hypothesis can be seamlessly integrated with any heterophilic GNNs designs, showcasing exceptional versatility. Whether it is for non-local neighbor extension or GNN architecture refinement, the snowflake can easily be incorporated as a plugin and demonstrates strong compatibility. (2) \textbf{Enhanced pruning requirement.}  In heterophilic graphs, given the higher probability of central nodes having different labels than surrounding nodes, the need for pruning aggregation channels becomes even more critical than in homogeneous graphs. This pruning aids nodes in selectively aggregating information and updating themselves effectively.

However, the original SnoH and its implementations appear to exhibit limitations when applied to heterophilic graphs. Firstly, SnoH relies on either \textit{gradient information} or \textit{cosine similarity} comparison for node early stopping. These heterophily-unaware approaches do not adequately integrate into heterophilic scenarios. Secondly, SnoH typically demonstrates convincing performance only in deep GNNs, which contradicts the prevailing focus on shallow designs for heterophilic GNNs. These necessitates bespoke strategies for heterophilic graphs. To handle the distinct characteristics between homophilic and heterophilic graphs, we introduce a \textbf{Heterophily-aware Early Stopping (HES)} strategy. HES benefits from two key aspects, thereby overcoming the limitations of SonH:

\begin{itemize}[leftmargin=*]
    \item[$\blacktriangleright$] HES employs a proxy label predictor, generating pseudo-label probability distributions for different nodes \cite{dai2022label, hu2021simple}. In this way, we can scrutinize the probability distributions of two connected nodes and determine the probabilities where two nodes are predicted to have the same label. This value can further represent the homophily score between two nodes, subsequently serving as a replacement for the original adjacent matrix.  

    \item[$\blacktriangleright$] By analyzing the variation in the heterophilic ratio across each layer, we are able to appropriately determine the depth for early stopping. This approach contrasts with SonH, which primarily exhibits efficacy in deeper GNN configurations.

\end{itemize}

\noindent Subsequently, by determining whether the heterophily of nodes increases before and after aggregation, we implement early stopping at the node receptive field level, thereby ensuring the efficacy of information aggregation. \textbf{For the first time}, we validate the existence of ``snowflakes'' in heterophilic graphs, underscoring the significance of the ``one node, one receptive field" paradigm in such scenarios. We introduce a universal solution to this issue, which stands out from previous designs due to its generality and model-agnostic nature. Remarkably, HES can \textbf{seamlessly} integrate with virtually \textbf{arbitrary} heterophilic designs, enhancing both its training and inference speeds. We summarize our contributions as follows:


\vspace{-0.2em}
\begin{itemize}[leftmargin=*]
\item We conceptualize ``one node one receptive field'' as heterophilic snowflake hypothesis in the heterophilic graph scenario. To achieve this target, we develop the heterophily-aware early stopping strategy, and offer theoretical analysis from the graph neural tangent kernel (GNTK) and stochastic block model (SBM) perspective to provide high-level insights of ours paradigm.


\item Hetero-S finds broad applicability and HES can aid various backbones in discovering optimal receptive fields for each node across diverse datasets. We verify HES on 10 GNN backbones across over 16 graph benchmarks. Experiments demonstrate that for all prevailing backbones, HES can facilitate substantial performance enhancements, ranging from 0.34\% $\sim$ 31.86\% in homophilic settings and from 0.68\% $\sim$ 21.73\% in heterophilic settings.


\item Similar to the conventional snowflake hypothesis, HES can scale up to deep GNNs effectively without any bells and whistles. Concretely, HES mitigates the performance degradation caused by excessive aggregation of heterophilic information, resulting in performance improvements ranging from 0.46\% $\sim$ 10.37\% in 16-layer configurations and from 1.13\% $\sim$ 7.92\% in 32-layer configurations. These experimental results demonstrate the potential of HES to be extended to large and densely connected graphs.

\item \textbf{More observations.} (I) Hetero-S has been empirically observed to exhibit comparable or even superior performance to the original snowflake hypothesis on both homophilic and heterophilic settings. In particular, Hetero-S demonstrates performance improvements ranging from 0.51\% $\sim$ 8.44\% on MixHop and JKNet in comparison to SnoH. (II) The \textbf{snowflakes (\textcolor{snowblue}{\ding{100}})} achieves the highest multiply-accumulate operations (MACs) saving ($25\% \sim 45\%$ of the baseline) compared to SOTA graph pruning algorithms~\cite{chen2021unified,liu2023dspar}, without any performance compromise.


\end{itemize}


\section{Preliminaries}

\subsection{Notations}
We consider an attributed graph denoted as $\mathcal{G} = \{\mathcal{V}, \mathcal{E}\}$, where $\mathcal{V}$ and $\mathcal{E}$ represent the sets of nodes and edges, respectively. The feature matrix of $\mathcal{G}$ is denoted as $\mathbf{X} \in \mathbb{R}^{N \times D}$, where $N = |\mathcal{V}|$ is the number of nodes in the graph and $D$ is the dimension of node features. We use $\mathbf{x}_i = \mathbf{X}[i,\cdot]$ to denote the $D$-dimensional feature vector corresponding to node $v_i \in \mathcal{V}$. An adjacency matrix $\mathbf{A} \in \mathbb{R}^{N \times N}$, serves to represent the connections between nodes, where $\mathbf{A}[i,j] = 1$ if $(v_i, v_j) \in \mathcal{E}$ and $0$ otherwise. To learn the node representations in a graph $\mathcal{G}$, most GNNs adhere to the following paradigm of neighborhood aggregation and message passing:

\vspace{-0.3em}
\begin{equation}
\mathbf{h}_i^{(l)} = \text{\fontfamily{lmtt}\selectfont \textbf{COMB}}\left( \mathbf{h}_i^{(l-1)}, \text{\fontfamily{lmtt}\selectfont \textbf{AGGR}}\{  \mathbf{h}_j^{(k-1)}: v_j \in \mathcal{N}(v_i) \} \right),\;0\leq l \leq L
\end{equation}
where $L$ is the number of GNN layers, $\mathbf{h}_i^{(0)} = \mathbf{x}_i$ denotes the feature vector of $v_i$ and $\mathbf{h}_i^{(l)} (1\leq l\leq L)$ denotes the node embedding of $v_i$ at the $l$-th GNN layer. {\fontfamily{lmtt}\selectfont \textbf{AGGR}} and {\fontfamily{lmtt}\selectfont \textbf{COMB}} represent functions used for aggregating neighborhood information and combining ego- and neighbor-representations, respectively. In the general node classification setting, after the graph convolution operations, GNN uses a linear mapping to map the node embedding $\mathbf{h}_i^{(l)}$ to the corresponding prediction probability value $\mathbf{z}_i$, and eventually get the model prediction $\hat{y}_i$:
\begin{equation}
\mathbf{z}_i = \operatorname{softmax}(\mathbf{h}_i^{(l)} \mathbf{W}),\; \text{and}\;\hat{y}_i=\arg\max\{\mathbf{z}_i\},
\end{equation}
where $\mathbf{W}$ is a learnable matrix, $\mathbf{z}_i \in \mathbb{R}^{C}$ signifies the probabilities of categorizing $v_i$ into each of the $C$ categories.

\subsection{Heterophily in Graph Neural Networks}

Following the previous heterophilic GNNs \cite{pei2020geom}, we define the node- and graph-level homophily ratio as follows:
\begin{equation}
    \text{node-level:} \quad \mathcal{H}^{(i)}_{\text{node}} = \frac{|\{v_j| v_j \in \mathcal{N}(v_i), \;y_i = y_j\}|}{|\mathcal{N}(v_i)|}  ,
\end{equation}
\begin{equation}
   \text{graph-level:} \quad \mathcal{H}_{\text{node}} = \frac{1}{|{\mathcal V}|} \sum_{v_i \in {\mathcal V}} \mathcal{H}^{(i)}_{\text{node}},
\end{equation}
where $\mathcal{N}(v_i)$ denotes the 1-hop neighbor of $v_i$ and $\mathcal{V}(v_i)$ denotes the 1-hop neighbors of $v_i$. Specifically, $\mathcal{H}^{(i)}_{\text{node}}$ represents the average proportion of neighbors that share the same class with node $v_i$, and $\mathcal{H}_{\text{node}}$ represents the global homophily by computing the average of node homophily. Conversely, the heterophily ratio at the node and graph level can be expressed as $\widetilde{\mathcal{H}}_{\text{node}}^{(i)} = 1 - \mathcal{H}_{\text{node}}^{(i)}$ and $\widetilde{\mathcal{H}}_{\text{node}} = 1 - \mathcal{H}_{\text{node}}$. In general, graphs that exhibit strong homophily tend to have $\mathcal{H}_{\text{node}}$ values approaching 1, whereas those characterized by pronounced heterophily often have values near 0.

It's essential to note that $\mathcal{H}_{\text{node}}^{(i)}$ solely reflects the heterophily within the 1-hop neighborhood of $v_i$. Furthermore, we extend this definition to the k-hop neighborhood:

\vspace{-0.3em}
\begin{equation}
\mathcal{NH}_i^k = \frac{|\{v_j| v_j \in \mathcal{N}^{(k)}(v_i), \;y_i = y_j\}|}{|\mathcal{N}^{(k)}(v_i)|} ,
\end{equation}
where $\mathcal{N}^{(k)}(v_i) = \{v_j| 1 \leq \text{ShortestPath}(v_j,v_i) \leq k\}$  represents the k-hop neighborhood of $v_i$.

\section{Motivation}\label{sec:motivation}

\begin{figure}[!ht]
  \centering
  \includegraphics[width=0.95\linewidth]{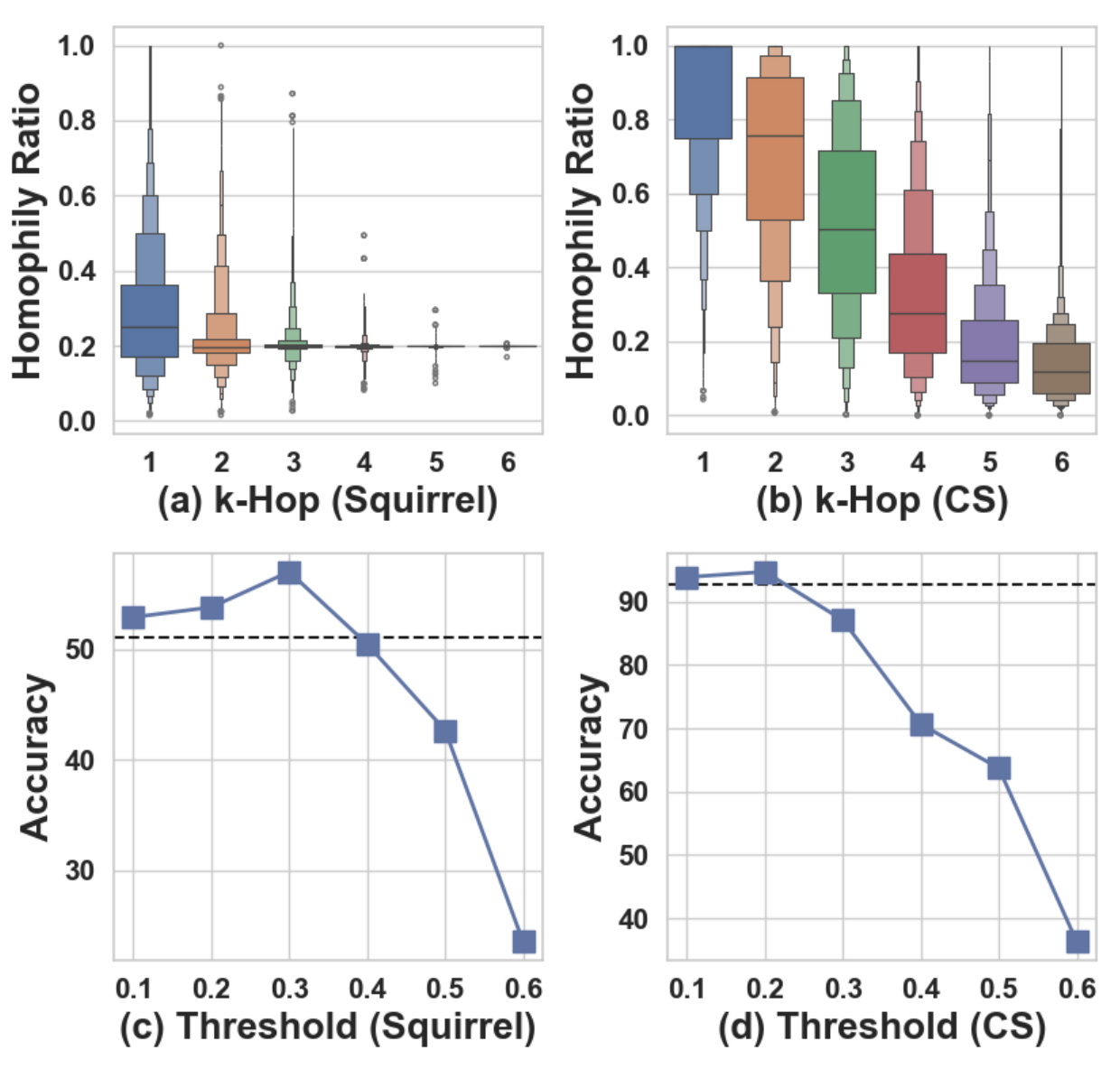}
  \vspace{-1.2em}
  \caption{The algorithm workflow of Heterophilic Snowflake Hypothesis (Hetero-S) and Heterophily-aware Early Stopping (HES). }
  \label{fig:motivation}
  \vspace{-0.5em}
\end{figure}

In this section, we prudently introspect the heterophily in different graphs and put forward the motivation of HES strategy. We start from the empirical observations. Concretely, we select both a heterophilic graph (Squirrel) and a homophilic graph (CS), and compute $\mathcal{NH}^k (1 \leq k \leq 6)$ for each node. As shown in  Figure~\ref{fig:motivation} (a,b), we list the following two observations: \textbf{Obs.1}: The homophily ratio of nodes in Squirrel significantly declines as the hop increases at a faster rate compared to those in CS; \textbf{Obs.2}: In both types of graphs, the distribution of node homophily ratios is diverse. Even when the receptive field size extends to 6 layers (\textit{i.e.}, 6-hop neighborhood), there exist nodes with homophily ratios approaching 1.

\noindent \textbf{Insights \& Reflections.} These observations naturally align with the concept of ``one node one receptive field." In both types of graph data, the k-hop homophily distributions of different nodes exhibit significant variation. This prompts the question: \textit{Is it feasible to determine the appropriate receptive field for each node based on its k-hop homophily distribution?} Going beyond this insight, we conduct the \textbf{empirical experiments} on Squirrel and CS with a 6-layer GNN. We determine their receptive field as follows:

\vspace{-0.5em}
\begin{equation}
\mathcal{R}_i = \underset{1 \leq j \leq k}{\operatorname{max}}(j \cdot \mathbbm{1}[{\mathcal{NH}^j_i \leq \epsilon}]),
\end{equation}
where $\mathcal{R}_i$ denotes the determined receptive filed for $v_i$ and $\epsilon$ is a pre-defined homophily threshold. After obtaining the receptive field, we then reformulate the aggregation operations as follows:
\vspace{-0.5em}
\begin{equation}\small
    \mathbf{h}_i^{(l)} = \left\{
    \begin{aligned}
        & \text{\fontfamily{lmtt}\selectfont \textbf{COMB}}\left( \mathbf{h}_i^{(l-1)}, \text{\fontfamily{lmtt}\selectfont \textbf{AGGR}}\{  \mathbf{h}_j^{(k-1)}: v_j \in \mathcal{N}(v_i) \} \right),\;0\leq l < \mathcal{R}_i;\\
        & \text{\fontfamily{lmtt}\selectfont \textbf{COMB}} \left(\mathbf{h}_i^{(l-1)}, \emptyset \right),\;\mathcal{R}_i \leq l \leq L
    \end{aligned}
    \right.    
\end{equation}

\noindent It is noteworthy that, beyond the $\mathcal{R}_i$-th layer of GNN, node $v_i$ ceases to aggregate information from its neighbors, thus achieving early stopping of the receptive field. Further, Figure~\ref{fig:motivation} (c,d) demonstrates the performance of GCN on Squirrel and CS with different homophily thresholds, which characterizes a $2.12\% \uparrow$ under $\epsilon=0.3$ on Squirrel, and a $1.34\% \uparrow$ under $\epsilon=0.2$ on CS. This confirms that early stopping of node receptive fields can indeed assist GNNs in learning more refined node representations. 
\begin{figure*}[t]
\setlength{\abovecaptionskip}{6pt}
\centering
\includegraphics[width=0.95\textwidth]{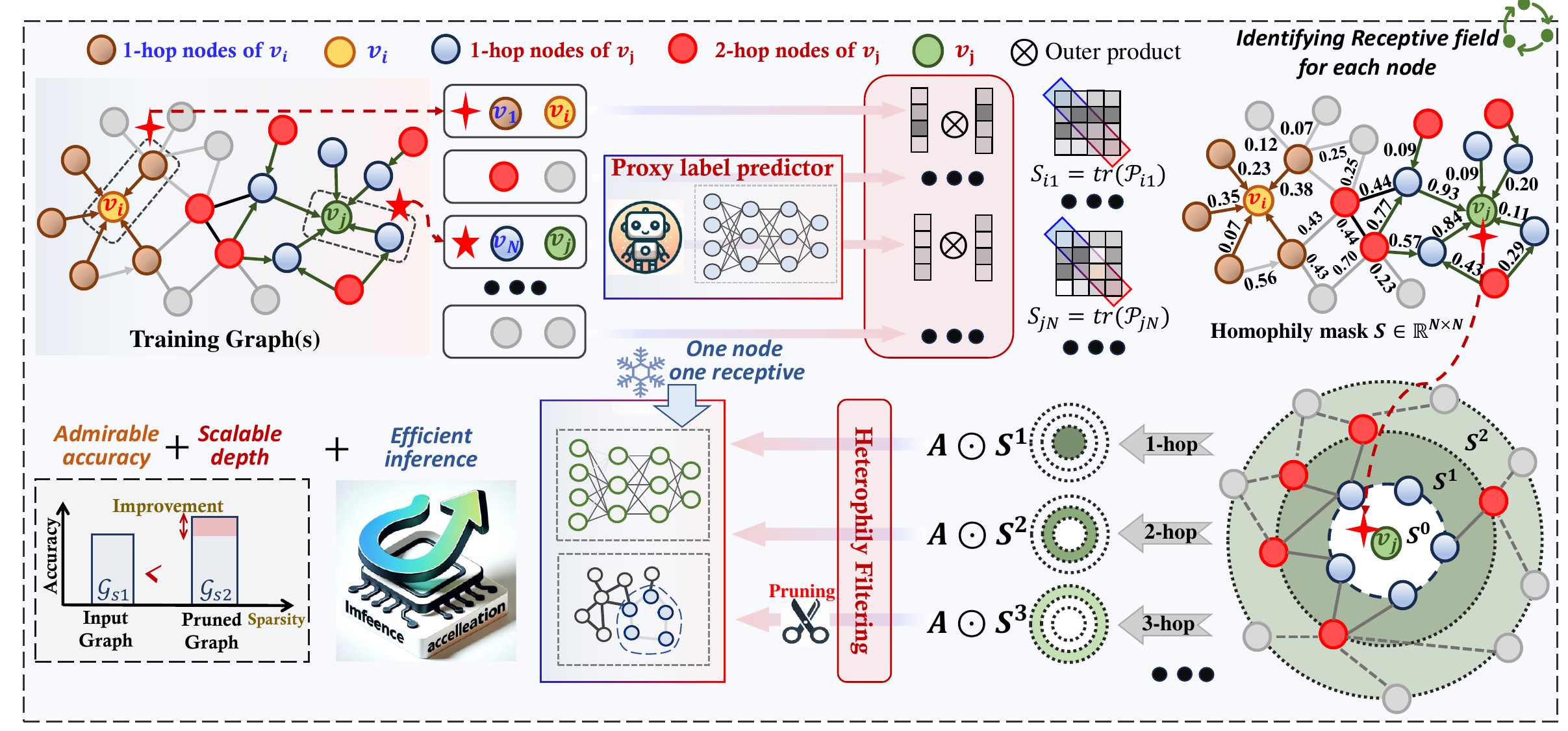}
\caption{The pipeline of our HES framework. For each node, we utilize a proxy model to evaluate the homophily strength of its edges, which is further used to estimate its multi-hop homophily ratio. Based on the homophily strength at each hop, we perform receptive field-level early stopping to determine a unique receptive field for each node.} \label{fig:method}
\end{figure*}

\section{Methodology}\label{sec:method}

Considering the aforementioned observation and gained motivations, we aim to identify an appropriate receptive field for each node. Analogous to the conventional snowflake hypothesis, we introduce the ``Heterophily Snowflake Hypothesis (Hetero-S)" for the first time in heterophilic graphs settings:

\vspace{0.4em}
\noindent\fbox{%
    \parbox{\linewidth}{%
        \textbf{Heterophilic Snowflake Hypothesis (Hetero-S)}: \textit{For an $L$-layer GNN on a heterophilic graph, each node possesses an optimal receptive field; training the GNN by aggregating information only from neighbors within this optimal receptive field minimizes the inclusion of excessive heterophilic information, yielding optimal representations. }
    }%
}
\vspace{0.2em}

\noindent More formally, consider an $L$-layer GNN, when optimized with stochastic gradient descent (SGD) on heterophilic graphs $\mathcal G = \{\mathbf{A},\mathbf{X}\}$, the GNN reaches a minimum validation loss and obtains a test accuracy of $\varphi$. Furthermore, let's assume the existence of a pruning algorithm guided by the $\mathcal{H}_{\text{node}}$, which ensures that each node $v_i$ has a unique optimal receptive field size of $k_i (1 \leq k_i \leq L)$, allowing it to be early stopped at the $k_i$-th GNN layer, \textit{i.e.}, set $\mathcal{N}^{(k+1)}(v_i) = \cdots = \mathcal{N}^{(L)}(v_i)=\emptyset$. This approach helps nodes avoid over-aggregation of heterophilic information, leading to a test accuracy of $\varphi'$. The Hetero-S posits that there exists optimal $k_i$ for each $v_i$ to satisfy that $\varphi' > \varphi$ (note as {\textcolor{snowblue}{\ding{100}}}).

\noindent To achieve this target, we can naturally resort to the implementation in Sec.~\ref{sec:motivation}. However, in practical scenarios (\textit{e.g.}, semi-supervised settings), a question arises regarding receptive field operation: \textit{when only partial nodes have known labels, how can we estimate the homophily ratio for each node within a $k$-hop neighborhood?} 




\subsection{Proxy Label Predictor}
As depicted above, in semi-supervised scenarios, we only know a small fraction of labels in a heterophilic graph. 
Towards this end, we formulate a GNN prediction process as a combination of two processes, \textit{i.e.}, $Y = \text{\fontfamily{lmtt}\selectfont \{\textbf{AGGR}} \diamond\text{\fontfamily{lmtt}\selectfont\textbf{COMB}}\} \diamond f_{Y}$. After the aggregation and combination process, $\text{\fontfamily{lmtt}\selectfont \{\textbf{AGGR}} \diamond\text{\fontfamily{lmtt}\selectfont\textbf{COMB}}\}$, nodes are mapped into a latent space characterized by distinguishability. This mapping ensures that nodes with the same label are positioned in similar locations, facilitating the identification of their relational patterns. Moreover, by employing an oracle function, we can effectively predict the outcome $Y$, leveraging the structured information encapsulated in this latent space. This process enhances the model's ability to discern and categorize nodes, significantly improving the accuracy and efficiency of our predictions. Upon reviewing previous work \cite{ma2021improving}, we first present a lemma:

\begin{lemma}
Assuming that $\mathcal{NH}^k_i$ for $v_i$ decreases w.r.t $k$ in proportion to $\zeta$ ($\zeta>1$), meaning that as the receptive field expands, $v_i$ aggregates more heterophilic information. Under such circumstances, when employing receptive field stopping, there exists $k \geq 2$ satisfying the condition that $\varphi( \{AGGR \diamond COMB\}^{(k)} \diamond f_{Y}) > \varphi( \{AGGR \diamond COMB\}^{(k + \pi)} \diamond f_{Y})$, where $\varphi(\cdot)$ is the performance metric, $\{AGGR \diamond COMB\}^{(k)}$ signifies aggregating information solely from the $k$-hop neighborhood, and $\pi \in \mathbb{N}^+$.
\end{lemma}

\noindent With this in mind, we construct a proxy label predictor ${{\mathcal P}_Y}$ to determine label-wise graph aggregation \cite{hu2021simple, dai2022label}. Concretely, we resort to a simple predictive model (here we can use MLP and GNN, and we place ablation results in Sec.~\ref{ablation}) to obtain pseudo probability label ${\tilde{\mathbf{z}_i}}$ with cross-entropy loss:

\vspace{-0.3em}
\begin{equation}
    \mathop {\min }\limits_{{\rm{\Theta }}} {\mathcal L}_{\text{CE}}\left( {{\mathcal G}_{tr}},{{\rm{\Theta }}} \right) =  - \frac{1}{{|{{\mathcal V}_{tr}}|}}\mathop \sum \limits_{{v_i} \in {{\mathcal V}_{tr}}} {\tilde{\boldsymbol{z}_i}}{\rm{log}}\left( {{\boldsymbol{z}_i}} \right),
    \label{eq:proxy}
\end{equation}
\vspace{-0.3em}

\noindent where ${{\mathcal G}_{tr}}$ denotes the training nodes (${{\rm{\mathcal V }}_{tr}}$) and graph structure, and ${\Theta}$ denotes network parameters. Different from previous work \cite{hu2021simple, dai2022label, wang2023heterophily}, we derive a pseudo probability distribution solely to determine the appropriate size of the receptive field.

\subsection{Training Homophily Mask}\label{sec:hes_overview}

After obtaining the node soft labels, we proceed to define the \textit{homophily mask} denoted as $\mathbf{S}$, where $\mathbf{S}_{ij}$, for edge $(i,j) \in \mathcal{E}$, represents the homophily strength of edge $(i,j)$,\textit{i.e.}, the likelihood that $v_i$ and $v_j$ share the same label. 

To obtain the expression for ${\mathbf{S}_{ij}}$, we calculate the label distribution for nodes ${v_i}$ and ${v_j}$ by computing the outer product of their respective predicted probability distributions. 

\begin{equation}\footnotesize
 S_{ij} = tr\left( {{\overset{\sim}{\boldsymbol{z}}}_{i}\otimes{\overset{\sim}{\boldsymbol{z}}}_{j}} \right) = tr\left(\begin{bmatrix}
{\overset{\sim}{\boldsymbol{z}}}_{1,1} & \cdots & {\overset{\sim}{\boldsymbol{z}}}_{1,C} \\
 \vdots & \ddots & \vdots \\
{\overset{\sim}{\boldsymbol{z}}}_{C,1} & \cdots & {\overset{\sim}{\boldsymbol{z}}}_{C,C}
\end{bmatrix} \right)= {\sum\limits_{\mu = 1\rightarrow C}{\overset{\sim}{\boldsymbol{z}}}_{\mu,\mu}}
\end{equation}

where ${\tilde z_{\alpha ,\beta }}$ denotes the probability that node ${v_i}$ belong to category $\alpha $, while node ${v_j}$ is associated with category $\beta $. $\otimes$ represents the out-product. Then we can formulate the $S$ via following equation:

\begin{equation}\footnotesize
    \mathbf{S} = \left( {\begin{array}{*{20}{c}}
{{S_{1,1}}}& \cdots &{{S_{1,N}}}\\
 \vdots & \ddots & \vdots \\
{{S_{N,1}}}& \cdots &{{S_{N,N}}}
\end{array}} \right)
\end{equation}

Note that $\mathbf{S}$ is parameterized by the proxy model ${{\mathcal P}_Y}$. The shapes of $\mathbf{S}$ and $\mathbf{A}$ are identical, in which we can co-optimize the weight ${\rm{\Theta }}$ and the $\mathbf{S}$ from end to end by utilizing objective ${{\mathcal L}_{\text{opt}}}$:

\begin{equation}\footnotesize
    {{\mathcal L}_{\text{opt}}} = {{\mathcal L}_{\text{CE}}}\left( {\left\{ {\mathbf{S} \odot \mathbf{A},\;\mathbf{X}} \right\},{\rm{\Theta }}} \right) = {{\mathcal L}_{\text{CE}}}\left( {\left\{ {{\rm{op}}{{\rm{t}}_S}\left\{ {\mathop {\min }\limits_{{{\rm{\Theta }}_{tr}}} {_{{{\mathcal G}_{tr}}}}\left( {{{\mathcal P}_Y}} \right)} \right\} \odot \mathbf{A},\;\mathbf{X}} \right\};{\rm{\Theta }}} \right)\label{eq:opt}
\end{equation}

\noindent where ${{\mathcal L}_{\text{CE}}}$ denotes cross-entropy loss, we optimize (${\rm{opt}}$) $\mathbf{S}$ by function ${{\rm opt}}_{S}\left\{ {\min\limits_{\Theta_{\mathit{tr}}}\mathbb{E}_{\mathcal{G}_{tr}}}\left( \mathcal{R}_{Y} \right) \right\}$ with minimized empirical risk $\mathbb{E}$.

\subsection{Early Stopping Based on Hop Heterophily} \label{sec:early_stopping}

After multiple rounds of training and optimization with Equation \ref{eq:opt}, we obtain a relatively accurate mask, denoted as $\dot{ \mathbf{S}}$. In the context of GNNs, multi-layer aggregation assists the model in capturing neighbor relationships at varying distances. Here, we employ $\dot{ \mathbf{S}}$ to replace $\mathbf{A}$ for capturing more distant neighbor relationships. Concretely, we use ${\dot{\mathbf{S}}^{(k)}} = {\dot{\mathbf{S}}^k} \in \mathbb{R}^{N \times N}$ to further represent the $k$-hop neighbor aggregation expression of homophily ratio. Subsequently, we calculate the row sum values for each hop, excluding self-loops:

\vspace{-0.5em}
\begin{equation}
    set\left\{ {\mathop \sum \nolimits_{\text{row}  \setminus \rm{diag}} {{\dot{\mathbf{S}}}^{\left( k \right)}}} \right\} = \{ \dot{\mathbf{S}}_1^{\left( k \right)},\dot{\mathbf{S}}_2^{\left( k \right)} \ldots \dot{\mathbf{S}}_N^{\left( k \right)} \},
\end{equation}
\noindent where $\dot{\mathbf{S}}_i^{(k)}$ represents the normalized row sum of the \( i \)-th row in $\dot{\mathbf{S}}^{(k)}$, where the summation excludes the self-loop represented by \( \text{row}\{k,k\} \) in the \( k \)-th row of the matrix.

\vspace{-0.5em}
\begin{equation} 
   \begin{aligned} 
      & \text{if} \;\;\; \dot S_i^{\left( k \right)} > \rho \dot S_i^{\left( {ES} \right)} \\
      & \text{then} \;\;\; {\mathcal D}o\left( {\mathcal{N}^{(ES)}(v_i) = \mathcal{N}^{(ES+1)}(v_i) =  \cdots   = \emptyset } \right),
   \end{aligned}
\end{equation}
\vspace{-0.3em}

\noindent where $\rho$  denotes the filtering threshold. While it is possible to assign a distinct threshold for each node, for simplicity, we employ a uniform $\rho$ across all nodes to filter the receptive field (the sensitivity analysis on $\rho$ is placed in Appendix \ref{sec:hyper}). ${\mathcal D}o$ denotes an intervention that forcefully assigns an aggregation status. We employ heterophily-aware early stopping at $ES$-th layer, which can ensure that each node possesses a unique receptive field size.

\subsection{Theoretical Analysis}

We provide a theoretical guarantee for heterophily snowflake hypothesis through graph neural tangent kernel (GNTK) \cite{du2019graph,huang2021towards}. Generally, during training, the NTK is deterministic and static \cite{jacot2018neural}, thus GNTK can be used to describe the training behavior of an infinitely-wide graph neural network.  In this work, we use GNTK to study the \textit{training dynamics} of an infinitely-wide GNN with node classification task. We provide the definition of GNKT as follows:
\begin{align} \label{eq:GNTK}
    K(u,u') = \sum_{l=1}^L  \frac{\partial f(u)}{\partial \theta^{(l)}} \frac{\partial f(u')}{\partial \theta^{(l)}} \triangleq \sum_{l=1}^L  K(u,u')^{(l)}
\end{align}
where $u$ and $u'$ are indexes of nodes in the graph, and $\theta^{(l)}$ is the set of all trainable parameters in $l$-th layer. According to Equation (\ref{eq:GNTK}), we need to calculate the GNTK at each layer, which follows the recursive relation: $\mathbf{K}^{(l)} = \mathbf{G}^{(l)} \mathbf{K}^{(l-1)}$, where $\mathbf{G}^{(l)} = \hat{\mathbf{A}}^{(l)} \Gamma( \hat{\mathbf{A}}^{(l)} )$ is $l$-th layer propagation matrix for the GNTK with $\hat{\mathbf{A}} = \mathbf{D}^{-1} \mathbf{A}$, $\Gamma$ denotes the transpose operation. Our target is to check if the smallest eigenvalue of GNTK is greater than zero or not. \textbf{According to \cite{hu2021simple}, when the smallest eigenvalue of GNTK is greater than zero, the training loss can be minimized to zero, implying successful optimization.} Inversely, if the smallest eigenvalue of is zero, then we would say that the GNN cannot be trained successfully. To study the eigenvalue of GNTK, we introduce a generative model named Stochastic Block Model (SBM) \cite{abbe2017community, funke2019stochastic}, which has been used in the theoretical analysis of GNNs. Then the following Lemma gives the smallest eigenvalue of the GNTK:

\vspace{-0.5em}
\begin{lemma}
Consider $\mathbf{A}$ is a probability adjacency matrix sampled from a SBM $\psi(N,p,q)$, wherein it is postulated that there are various clusters interconnected with an internal connection probability of $p$, and the inter-cluster connectivity rate is $q$. We can conclude the expected smallest eigenvalue of the GNTK is given by $\mathbb{E}_{\mathbf{A} \sim \psi(N,p,q)} [\lambda] = {\rm{\Pi }}_{i = 1}^L \frac{1 - p}{\left( N - 1 \right)p + Nq + 1}$.
\end{lemma}

In SBM, $p$ and $q$ can be deemed as the expectation of the adjacency matrix. With this in mind, we observe the $\mathbf{K}^{(l)} =  \mathbf{G}^{(l)} \mathbf{K}^{(l-1)}= \cdots = \mathbf{G}^{(l)}\cdots \mathbf{G}^{(1)} \mathbf{K}^{(0)}$. Subsequently, we can express the general probabilistic form of $\mathbf{K}^{(l)}$:

\vspace{-0.3cm}
\begin{align} \footnotesize
\label{eq:16}
    \mathbf{K}^{(l)} = \hat{\mathbf{A}}^{(l)} \Gamma( \hat{\mathbf{A}}^{( l )} ) \hat{\mathbf{A}}^{(l-1)} \Gamma( \hat{\mathbf{A}}^{( l-1)} ) \cdots   \hat{\mathbf{A}}^{(1)} \Gamma( \hat{\mathbf{A}}^{(1)} ) \mathbf{K}^{(0)}
\end{align}
\vspace{-0.3cm}

\noindent where $l$-th adjacency matrix $\hat{\mathbf{A}}^{(l)}$ is further characterized in probabilistic form, and the density of  $\hat{\mathbf{A}}^{(l)}$ is correlated with the magnitude of $p^{(l)}$ and $q^{(l)}$. We observe that with the increasing depth of GNN, early stopping of the receptive field can assist in the hierarchical decrement of $p$ and $q$. Consider a simple binary classification scenario with a balanced distribution, where each category consists of $N$ nodes. The eigenvalues of matrix $\mathbb{E}_{\mathbf{A} \sim \psi(N,p,q)} [\mathbf{G}]$ are as follows:
\vspace{-0.3em}
\begin{equation} \small
    {\lambda _0} = 1,{\lambda _1} = \frac{{(N-1)p - Nq + 1}}{{(N-1)p + Nq + 1}},{\lambda _2} =  \cdots {\lambda _{2N-1}} = \frac{{1-p}}{{(N-1)p + Nq + 1}}
\end{equation}
Generally,  the probability of connection between nodes with the same class label is higher than that of different classes, \textit{i.e.}, $p > q$. Hence, the inequality $(1-p) < (N-1)p - Nq + 1$ is always satisfied. Therefore, the smallest eigenvalue of the SBM is given by $\frac{1-p}{(N-1)p + Nq + 1}$. Considering in subsequent layers, the HES algorithm is applied such that $p$ and $q$ remain attenuation while heterophilic nodes are pruned (the decay rate of $p$ is slower), the product of the smallest eigenvalues is given by:
\begin{equation}\footnotesize
    {\rm{\Pi }}_{i = 1}^L\frac{{1 - {p^{(i)}}}}{{\left( {N - 1} \right){p^{(i)}} + N{q^{(i)}} + 1}} = {\rm{\Pi }}_{i = 1}^L\left( {1 - \frac{{N\left( {{p^{(i)}} + {q^{\left( i \right)}}} \right)}}{{\left( {1 - {p^{(i)}}} \right) + N\left( {{p^{(i)}} + {q^{\left( i \right)}}} \right)}}} \right)
\end{equation}

\noindent Consider \( p > q \) and p,q are both functions of \( N \). Here, we conduct a case study where both $p$ and $q$ decay quadratically with $i$. Without loss of generality, we assume $p = 1/((N-1)*i^2)$ and $q = 1/(N*i^2)$. We observe that, under these conditions, the GNTK eventually diverges to $ \sqrt{2N} \operatorname{csch}(\sqrt{2} \pi) \sin(\pi/\sqrt{N}) $, which concludes the proof. See details in Appendix \ref{app:proof}.

\section{Experiments}
In this section,  we conduct extensive experiments to answer
the following research questions (\textbf{RQ}):

\setlength{\tabcolsep}{3pt}
\begin{table*}[t]
\centering
\caption{Qantitative performance for different layers, we report \underline{the average results of FIVE runs} and record the results after adding Hetero-S (+{\textcolor{snowblue}{\ding{100}}}). ``$8$\textcolor{snowblue}{\ding{100}}'' in \twemoji{crown} column denotes $8$-th layer GNN with HES shows the best performance.  \textbf{OOM} represents out-of-memory.}
\label{tab:oversmoothing}
\vspace{-0.30cm}
\begin{adjustbox}{width=1\linewidth}
{\footnotesize
\begin{tabular}{c c cccccc cccccc ccccc}
\toprule
    \textbf{Layers} 
        && \textbf{2}          
        & \multicolumn{1}{c}{\textbf{4}} 
        & \multicolumn{1}{c}{\textbf{6}} 
        & \multicolumn{1}{c}{\textbf{8}} 
        & \multicolumn{1}{c}{\textbf{\twemoji{crown}}}

        && \textbf{2} 
        & \multicolumn{1}{c}{\textbf{4}} 
        & \multicolumn{1}{c}{\textbf{6}} 
        & \multicolumn{1}{c}{\textbf{8}} 
        & \multicolumn{1}{c}{\textbf{\twemoji{crown}}}

        && \textbf{2} 
        & \multicolumn{1}{c}{\textbf{4}} 
        & \multicolumn{1}{c}{\textbf{6}} 
        & \multicolumn{1}{c}{\textbf{8}} 
        & \multicolumn{1}{c}{\textbf{\twemoji{crown}}}\\
        \cmidrule{1-1} \cmidrule{3-7} \cmidrule{9-13} \cmidrule{15-19} 
&& \multicolumn{5}{c}{\textbf{Cora} ($\mathcal{H}_{\text{node}}=0.57$)}     & & \multicolumn{5}{c}{\textbf{Citeseer} ($\mathcal{H}_{\text{node}}=0.74$)}  & & \multicolumn{5}{c}{\textbf{PubMed} ($\mathcal{H}_{\text{node}}=0.80$)}  \\ 
\cline{3-7} \cline{9-13} \cmidrule{15-19}
                                    
    \multicolumn{1}{l}{GCN/ +{\textcolor{snowblue}{\ding{100}}}}       
    &&$81.60$ & {$82.10/\mathbf{82.73}$}  & {$ 83.20/\mathbf{84.00}$} & {$83.90/\mathbf{85.10}$} & $8$\textcolor{snowblue}{\ding{100}}
    
    &&  {$ 72.80$}  & $73.20/\textbf{73.40}$ & $\textbf{73.80}/73.40$ & {$73.20/\mathbf{74.20}$} & $8$\textcolor{snowblue}{\ding{100}}
    
    && {$ 87.70$} & {$ 87.90/\mathbf{88.50}$} & {$ 87.40/\mathbf{88.12}$}    & $87.50/\mathbf{87.98}$ & $4$\textcolor{snowblue}{\ding{100}}\\ 
    \multicolumn{1}{l}{GIN/ +{\textcolor{snowblue}{\ding{100}}}}       
    &&$73.60$ & {$78.30/\mathbf{79.70}$}  & {$81.10/\mathbf{81.30}$} & {$78.80/\mathbf{82.10}$} & $8$\textcolor{snowblue}{\ding{100}}
    
    &&  {$ 60.51$}  & $68.47/\mathbf{72.96}$& $73.08/\mathbf{73.50}$ & {$72.60/\mathbf{74.50}$} & $8$\textcolor{snowblue}{\ding{100}}      
    && {$ 87.60$} & {$ 86.37/\mathbf{88.18}$} & {$ 88.51/\mathbf{89.37}$}    & $88.45/\mathbf{88.70}$ & {$6$}\textcolor{snowblue}{\ding{100}} \\ 
    \multicolumn{1}{l}{GAT/ +{\textcolor{snowblue}{\ding{100}}}}       
    &&$82.70$ & {$ 83.52/\mathbf{84.28}$}  & {$ 85.31/\mathbf{85.42}$} & {$ 83.40/\mathbf{85.19}$} & $6$\textcolor{snowblue}{\ding{100}}
    
    &&  {$73.78$}  & $75.15/\mathbf{75.98}$ & $74.21/\mathbf{74.30}$ & {$\mathbf{74.15}/73.80$} & $4$\textcolor{snowblue}{\ding{100}}      
    && {$ 87.40$} & {$ 85.12/\mathbf{85.52}$} & {$85.10/\mathbf{85.50}$}    & $84.32/\mathbf{85.08}$ & {$2$} \\ 
    
    \multicolumn{1}{l}{Mixhop/ +{\textcolor{snowblue}{\ding{100}}}}       
    &&$85.20$ 
    & {$82.80/\mathbf{84.07}$}  
    & {$82.10/\mathbf{82.80}$} 
    & {$82.16/\mathbf{83.90}$} 
    & {$2$} 
    &&  {$76.98$}  & $75.90/\mathbf{77.25}$ & $75.50/\mathbf{77.88}$ & {$74.30/\mathbf{77.10}$} & $6$\textcolor{snowblue}{\ding{100}}
    
    && {$77.20$} & {$77.30/\mathbf{77.40}$} & {$75.70/\mathbf{76.70}$}   & {$73.20/\mathbf{77.28}$}    & $4$\textcolor{snowblue}{\ding{100}} \\ 
    
    \multicolumn{1}{l}{Geom-GCN/ +{\textcolor{snowblue}{\ding{100}}}}       
    &&$85.35$ & {$ 21.02/\mathbf{86.77}$}  & {$19.42/\mathbf{81.24}$} & {$ 14.67/\mathbf{78.99}$} & $4$\textcolor{snowblue}{\ding{100}} 
    &&  {$78.42$}  & $29.98/\mathbf{67.43}$ & $25.66/\mathbf{68.47}$ & {$12.04/\mathbf{64.26}$} & 2     
    && {$85.95$} & {$ 73.22/\mathbf{87.03}$} & {$ 43.66/\mathbf{80.38}$} &$40.58/\mathbf{74.63}$   & {$4$}\textcolor{snowblue}{\ding{100}}  \\ 
    
    \multicolumn{1}{l}{H2GCN/ +{\textcolor{snowblue}{\ding{100}}}}       
    &&$77.05 $ & {$ 75.74/\mathbf{76.52}$}  & {$ 87.14/\mathbf{87.83}$} & {$ 85.76/\mathbf{87.98}$} & {$8$}\textcolor{snowblue}{\ding{100}}
    &&  {$ 78.02$}  & {$76.42/\mathbf{78.25}$} & {$75.64/\mathbf{78.18}$} & {$75.46/\mathbf{78.24}$} & {$8$}\textcolor{snowblue}{\ding{100}}      
    && {$ 88.54$} & {$ 87.74/\mathbf{88.66}$} & {$ 86.02 /\mathbf{88.72}$}    & {$85.28/\mathbf{88.52}$} & {$6$}\textcolor{snowblue}{\ding{100}} \\ 

    \multicolumn{1}{l}{GCNII/ +{\textcolor{snowblue}{\ding{100}}}}       
    &&$84.40$ & {$79.50/\mathbf{84.85}$}  & {$76.30/\mathbf{81.50}$} & {$47.90/73.60$} & {$4$}\textcolor{snowblue}{\ding{100}} 
    &&  {$74.60$}  & $\mathbf{73.32}/72.89$ & $71.70/\mathbf{72.66}$ & {$54.79/\mathbf{57.99}$} & 2       
    && {$ 86.50$} & {$ 85.57/\mathbf{87.25}$} & {$84.00/\mathbf{84.92}$}    & $82.53/\mathbf{84.60}$ & {$4$}\textcolor{snowblue}{\ding{100}} \\ 

    \multicolumn{1}{l}{GPNN/ +{\textcolor{snowblue}{\ding{100}}}}       
    &&$81.20$ & {$80.73/\mathbf{82.80}$}  & {$43.70/\mathbf{78.99}$} & {$46.49/\mathbf{62.18}$} & {$4$}\textcolor{snowblue}{\ding{100}}
    
    &&  {$74.20$}  & $73.30/\mathbf{74.81}$& $73.80/\mathbf{74.38}$ & {$72.50/\mathbf{73.64}$} &{$4$}\textcolor{snowblue}{\ding{100}}      
    && {$88.66$} & {$ 89.60/\mathbf{89.71}$} & {$40.70/\mathbf{88.19}$}    & $40.70/\mathbf{85.10}$ & {$4$}\textcolor{snowblue}{\ding{100}} \\ 

    \multicolumn{1}{l}{JKNet/ +{\textcolor{snowblue}{\ding{100}}}}       
    &&$83.00$ & {$ 82.51/\mathbf{83.37}$}  & {$80.50/\mathbf{83.19}$} & {$81.60/\mathbf{83.39}$} & {$4$}\textcolor{snowblue}{\ding{100}} 
    
    &&  {$ 74.19$}  & $73.28/\mathbf{74.42}$& $72.60/\mathbf{73.88}$ & {$73.10/\mathbf{74.05}$} & {$4$}\textcolor{snowblue}{\ding{100}}        
    && {$88.78$} & {$ 88.20/\mathbf{89.41}$} & {$88.78/\mathbf{88.06}$}    & $87.70/\mathbf{88.90}$ &{$4$}\textcolor{snowblue}{\ding{100}}  \\ 

    \multicolumn{1}{l}{MGNN/ +{\textcolor{snowblue}{\ding{100}}}}       
    &&$66.80$ & {$63.60/\mathbf{74.72}$}  & {$73.81/\mathbf{75.27}$} & {$79.30/\mathbf{84.88}$} & {$8$}\textcolor{snowblue}{\ding{100}} 
    
    &&  {$43.90$}  & $53.30/\mathbf{70.46}$ & $69.50/\mathbf{78.33}$ & {$77.08/\mathbf{79.30}$} & {$8$}\textcolor{snowblue}{\ding{100}}        
    && {$89.20$} & {$ 90.07/\mathbf{90.20}$} & {$88.30/\mathbf{89.47}$}  & $88.27/\mathbf{89.16}$  & {$4$}\textcolor{snowblue}{\ding{100}}  \\ 
     
    \midrule 

&& \multicolumn{5}{c}{\textbf{Texas} ($\mathcal{H}_{\text{node}}=0.11$)}     & & \multicolumn{5}{c}{\textbf{Wisconsin} ($\mathcal{H}_{\text{node}}=0.21$)}  & & \multicolumn{5}{c}{\textbf{Cornell} ($\mathcal{H}_{\text{node}}=0.22$)}                                                                         \\ 
\cline{3-7} \cline{9-13} \cmidrule{15-19}
                                    
    \multicolumn{1}{l}{GCN/ +{\textcolor{snowblue}{\ding{100}}}}       
    &&$68.42$ & {$73.68/\mathbf{78.95}$}  & {$68.42/\mathbf{71.05}$} & {$60.53/\mathbf{63.16}$} & {$4$}\textcolor{snowblue}{\ding{100}}
    &&  {$56.86$}  & 
    
    {$41.48/\mathbf{50.98}$}& $47.06/\mathbf{60.78}$ & {$52.94/\mathbf{54.90}$} & {$6$}\textcolor{snowblue}{\ding{100}} 
    && {$36.84$} & {$34.21/\mathbf{44.74}$} & {$39.47/\mathbf{47.37}$}    & $34.21/\mathbf{39.47}$ & {$6$}{\textcolor{snowblue}{\ding{100}}} \\ 
    \multicolumn{1}{l}{GIN/ +{\textcolor{snowblue}{\ding{100}}}}       
    &&$63.16$ & {$71.02/\mathbf{76.32}$}  & {$ 73.68/\mathbf{76.32}$} & {$ 77.84/\mathbf{78.85}$} & {$8$}\textcolor{snowblue}{\ding{100}}
    
    &&  {$ 62.75$}  & $45.10/\mathbf{54.08}$ & $54.88/\mathbf{58.82}$ & {$52.17/\mathbf{62.84}$} & {$8$\textcolor{snowblue}{\ding{100}}}       
    && {$34.21$} & {$28.95/\mathbf{36.27}$} & {$26.77/\mathbf{38.42}$}    & $36.84/\mathbf{37.20}$ & {$6$\textcolor{snowblue}{\ding{100}}}  \\ 
    \multicolumn{1}{l}{GAT/ +{\textcolor{snowblue}{\ding{100}}}}       
    &&$60.53$ & {$60.53/\mathbf{65.79}$}  & {$ 57.89/\mathbf{65.79}$} & {$ 60.53/\mathbf{63.16}$} & {$6$\textcolor{snowblue}{\ding{100}}} 
    &&  {$52.94$}  & $51.79/\mathbf{54.90}$ & $49.62/\mathbf{56.86}$ & {$50.26/\mathbf{54.33}$} & {$6$\textcolor{snowblue}{\ding{100}}}     
    && {$36.84$} & {$ 31.58/\mathbf{39.76}$} & {$26.32/\mathbf{34.21}$}    & $26.38/\mathbf{36.84}$ & {$4$\textcolor{snowblue}{\ding{100}}} \\ 
    
    \multicolumn{1}{l}{Mixhop/ +{\textcolor{snowblue}{\ding{100}}}}       
    &&$92.11$ & {$ 89.47/\mathbf{94.86}$}  & {$ 86.44/\mathbf{89.68}$} & {$\mathbf{78.95}/76.32$} & {$4$\textcolor{snowblue}{\ding{100}}}
    &&  {$80.39$}  & $82.35/\mathbf{84.31}$ & $74.51/\mathbf{82.93}$ & {$68.63/\mathbf{80.28}$} & {$4$\textcolor{snowblue}{\ding{100}}}      
    && {$73.68$} & {$ 52.63/\mathbf{68.99}$} & {$50.80/\mathbf{64.32}$}    & $39.47/\mathbf{65.70}$ & 2\\ 
    
    \multicolumn{1}{l}{Geom-GCN/ +{\textcolor{snowblue}{\ding{100}}}}       
    &&$66.53$ & {$ 60.84/\mathbf{66.98}$}  & {$43.17/\mathbf{60.34}$} & {$43.08/\mathbf{58.26}$} & {$4$}\textcolor{snowblue}{\ding{100}} 
    &&  {$64.51$}  & $61.88/\mathbf{66.92}$ & $36.87/\mathbf{54.22}$ & {$36.87/\mathbf{55.92}$} & {$4$}\textcolor{snowblue}{\ding{100}}       
    && {$60.54$} & {$ 24.78/\mathbf{48.99}$} & {$ 24.78/\mathbf{52.96}$} &$24.78/\mathbf{46.71}$   & 2 \\ 
    
    \multicolumn{1}{l}{H2GCN/ +{\textcolor{snowblue}{\ding{100}}}}       
    &&$89.54$ & {$ 68.52/\mathbf{92.02}$}  & {$ 73.43/\mathbf{92.28}$} & {$ 73.78/\mathbf{89.26}$} & {$ 6$\textcolor{snowblue}{\ding{100}}} 
    &&  {$ 76.42$}  & {$72.75/\mathbf{78.23}$} & {$68.73/\mathbf{76.41}$} & {$ 72.48/\mathbf{78.62}$} & {$ 8$\textcolor{snowblue}{\ding{100}}}
    && {$ 55.76$} & {$ \mathbf{65.45}/63.52$} & {$ \mathbf{65.24}/63.02$}    & {$44.62/\mathbf{57.46}$} & {$ 4$\textcolor{snowblue}{\ding{100}}}\\ 

    \multicolumn{1}{l}{GCNII/ +{\textcolor{snowblue}{\ding{100}}}}       
    &&$72.68$ & {$71.65/\mathbf{73.68}$}  & {$65.79/\mathbf{68.73}$} & {$63.16/\mathbf{63.89}$} & {$4$} {\textcolor{snowblue}{\ding{100}}}
    &&  {$45.10$}  & $49.02/\mathbf{51.38}$ & $45.77/\mathbf{49.86}$ & {$42.71/\mathbf{45.10}$} & {$4$} {\textcolor{snowblue}{\ding{100}}}   
    && {$73.68$} & {$ 52.63/\mathbf{68.49}$} & {$ 50.00/\mathbf{69.72}$}    & $39.47/\mathbf{66.10}$ & 2 \\ 

    \multicolumn{1}{l}{GPNN/ +{\textcolor{snowblue}{\ding{100}}}}       
    &&$78.95$ & {$ 60.53/\mathbf{84.21}$}  & {$60.53/\mathbf{83.18}$} & {$73.66/\mathbf{78.06}$} & {$4$}\textcolor{snowblue}{\ding{100}} 
    
    &&  {$66.67$}  & $70.59/\mathbf{71.28}$ & $\mathbf{72.35}/70.16$ & {$68.39/\mathbf{76.91}$} & {$8$}\textcolor{snowblue}{\ding{100}}       
    && {$50.00$} & {$ 52.83/\mathbf{63.24}$} & {$\mathbf{50.06}/49.66$}    & $49.72/\mathbf{52.47}$ & {$4$}\textcolor{snowblue}{\ding{100}}   \\ 

    \multicolumn{1}{l}{JKNet/ +{\textcolor{snowblue}{\ding{100}}}}       
    &&$74.89$ & {$ 78.95/\mathbf{84.21}$}  & {$84.07/\mathbf{84.30}$} & {$84.07/\mathbf{89.88}$} & {$8$}\textcolor{snowblue}{\ding{100}} 
    &&  {$47.60$}  & $60.38/\mathbf{62.94}$ & $60.38/\mathbf{64.99}$ & {$56.17/\mathbf{63.82}$} & {$6$}\textcolor{snowblue}{\ding{100}}       
    && {$34.21$} & {$ 42.17/\mathbf{47.25}$} & {$43.66/\mathbf{50.07}$}    & $44.89/\mathbf{48.65}$ & {$6$}\textcolor{snowblue}{\ding{100}}  \\ 

    \multicolumn{1}{l}{MGNN/ +{\textcolor{snowblue}{\ding{100}}}}       
    &&$84.21$ & {$ 86.84/\mathbf{92.31}$}  & {$88.37/\mathbf{89.64}$} & {$84.21/\mathbf{93.09}$} & {$8$}\textcolor{snowblue}{\ding{100}}   
    &&  {$72.55$}  & $84.31/\mathbf{86.27}$& $82.35/\mathbf{88.77}$ & {$80.17/\mathbf{86.20}$} & {$6$}\textcolor{snowblue}{\ding{100}}       
    && {$55.26$} & {$ {68.18}/\mathbf{71.22}$} & {$\mathbf{65.99}/71.08$} & $65.03/\mathbf{70.83}$   & {$4$}\textcolor{snowblue}{\ding{100}}  \\ 
     
    \midrule

&& \multicolumn{5}{c}{\textbf{Squirrel} ($\mathcal{H}_{\text{node}}=0.22$)}     & & \multicolumn{5}{c}{\textbf{Chameleon} ($\mathcal{H}_{\text{node}}=0.23$)}  & & \multicolumn{5}{c}{\textbf{Actor} ($\mathcal{H}_{\text{node}}=0.22$)}                                                                         \\ 
\cline{3-7} \cline{9-13} \cmidrule{15-19}
                                    
    \multicolumn{1}{l}{GCN/ +{\textcolor{snowblue}{\ding{100}}}}       
    &&$55.83$ & {$53.51/\textbf{56.20}$}  & {$ 51.01/\textbf{51.87}$} & {$49.76/\textbf{53.74}$} & {$4${\textcolor{snowblue}{\ding{100}}}} 
    &&  {$67.11$}  & $\textbf{63.82}/63.39$ & $59.65/\textbf{63.38}$ & {$57.89/\textbf{60.31}$} & {$2$}       
    && {$29.98$} & {$\textbf{26.64}/25.67$} & {$27.50/\textbf{28.60}$}  & $26.83/\textbf{28.22}$  & 2 \\ 
    \multicolumn{1}{l}{GIN/ +{\textcolor{snowblue}{\ding{100}}}}       
    &&$46.69$ & {$43.26/\mathbf{47.93}$}  & {$ 43.23/\mathbf{46.89}$} & {$39.48/\mathbf{44.86}$} & {$4${\textcolor{snowblue}{\ding{100}}}} 
    &&  {$63.82$}  & $62.28/\mathbf{66.89}$ & $60.31/\mathbf{64.04}$ & {$58.11/\mathbf{61.62}$} & {$4$}\textcolor{snowblue}{\ding{100}}       
    && {$29.34$} & {$ 29.87/30.83$} & {$23.95/25.46$}    & $24.41/25.68$ & {$4$}\textcolor{snowblue}{\ding{100}}  \\ 
    \multicolumn{1}{l}{GAT/ +{\textcolor{snowblue}{\ding{100}}}}
    &&$60.42$ & {$46.11/\mathbf{47.55}$}  & {$19.31/\mathbf{28.41}$} & {$21.33/\mathbf{26.33}$} & 2
    &&  {$ 69.74$}  & $70.61/\mathbf{70.83}$ & $\mathbf{66.89}/66.23$ & {$61.84/\mathbf{63.77}$} & {$4$}\textcolor{snowblue}{\ding{100}}       
    && {$28.09$} & {$ 27.57/\mathbf{27.87}$} & {$25.26/\mathbf{25.86}$} & $25.13/\mathbf{26.58}$   & 2 \\ 
    
    \multicolumn{1}{l}{Mixhop/ +{\textcolor{snowblue}{\ding{100}}}}       
    &&$54.76$ & {$56.20/\mathbf{56.88}$}  & {$\mathbf{54.84}/53.67$} & {$52.55/\mathbf{52.87}$} & {$4$}\textcolor{snowblue}{\ding{100}} 
    
    &&  {$66.45$}  & $64.69/\mathbf{69,08}$ & $63.38/\mathbf{66.04}$ & {$62.67/\mathbf{65.89}$} & {$4$}\textcolor{snowblue}{\ding{100}}      
    && {$32.63$} & {$ 31.25/\mathbf{34.80}$} & {$35.72/\mathbf{36.68}$}  & $35.20/\mathbf{35.28}$  & {$6$}\textcolor{snowblue}{\ding{100}}  \\ 
    
    \multicolumn{1}{l}{Geom-GCN/ +{\textcolor{snowblue}{\ding{100}}}}       
    &&$38.44$ & {$ 36.47/\mathbf{40.33}$}  & {$31.42/\mathbf{38.67}$} & {$27.00/\mathbf{36.92}$} & {$4$}\textcolor{snowblue}{\ding{100}} 
    &&  {$60.75$}  & $61.42/\mathbf{63.92}$ & $55.73/\mathbf{59.12}$ & {$54.32/\mathbf{58.46}$} &{$4$}\textcolor{snowblue}{\ding{100}}        
    && {$31.59$} & {$ 22.64/\mathbf{32.26}$} & {$ 22.64/\mathbf{33.40}$}  &$22.64/\mathbf{28.21}$  & {$6$}\textcolor{snowblue}{\ding{100}}  \\ 
    
    \multicolumn{1}{l}{H2GCN/ +{\textcolor{snowblue}{\ding{100}}}}       
    &&$27.35$ & {$ \mathbf{27.94}/27.04$}  & {$ \mathbf{30.07}/28.21$} & {$ \textbf{OOM}$} & {$ 6$}\textcolor{snowblue}{\ding{100}}
    &&  {$ 55.86$}  & {$\mathbf{55.92}/53.82$} & {$53.07/\mathbf{55.24}$} & {$ 51.24/\mathbf{56.04}$} & {$8$}\textcolor{snowblue}{\ding{100}}       
    && {$ 33.24$} & {$ 32.66/\mathbf{33.74}$} & {$\mathbf{33.58}/33.35$}    & {$33.02/\mathbf{33.47}$} & {$4$}\textcolor{snowblue}{\ding{100}}\\

    \multicolumn{1}{l}{GCNII/ +{\textcolor{snowblue}{\ding{100}}}}       
    &&$40.61$ & {$37.94/\mathbf{43.99}$}  & {$\mathbf{33.24}/30.58$} & {$28.43/\mathbf{29.66}$} & 4{\textcolor{snowblue}{\ding{100}}} 
    &&  {$66.89$}  & $54.39/\mathbf{54.69}$ & $44.36/\mathbf{48.46}$ & {$\mathbf{46.49}/29.77$} & 2     
    && {$31.84$} & {$ 24.61/\mathbf{24.67}$} & {$24.93/\mathbf{26.85}$}    & $24.47/\mathbf{24.80}$ & 2\\ 

    \multicolumn{1}{l}{GPNN/ +{\textcolor{snowblue}{\ding{100}}}}       
    &&$43.04$ & {$ 27.95/\mathbf{38.74}$}  & {$18.54/\mathbf{29.47}$} & {$18.54/\mathbf{28.61}$} & 2 
    &&  {$64.47$}  & $50.66/\mathbf{65.55}$& {$48.03/\mathbf{52.77}$}& {$29.39/\mathbf{45.08}$} & {$4$}\textcolor{snowblue}{\ding{100}}       
    && {$24.67$} & {$ 25.00/\mathbf{25.63}$} & {$24.67/\mathbf{24.88}$}  & $22.18/\mathbf{25.09}$  & {$4$}\textcolor{snowblue}{\ding{100}}  \\ 

    \multicolumn{1}{l}{JKNet/ +{\textcolor{snowblue}{\ding{100}}}}       
    &&$47.36$ & {$ 59.75/\mathbf{61.08}$}  & {$ 58.69/\mathbf{61.22}$} & {$57.83/\mathbf{61.48}$} & {$8$}\textcolor{snowblue}{\ding{100}} 
    
    &&  {$64.25$}  & $70.18/\mathbf{72.37}$ & $\mathbf{70.83}/69.44$ & {$70.61/\mathbf{70.93}$} & {$4$}\textcolor{snowblue}{\ding{100}}       
    && {$ 28.09$} & {$ 27.17/\mathbf{29.87}$} & {$27.57/\mathbf{30.04}$}    & $28.09/\mathbf{29.65}$ & {$6$}\textcolor{snowblue}{\ding{100}}  \\ 

    \multicolumn{1}{l}{MGNN/ +{\textcolor{snowblue}{\ding{100}}}}       
    &&$41.79$ & {$45.44/\mathbf{48.76}$}  & {$41.31/\mathbf{43.79}$} & {$39.00/\mathbf{40.28}$} & {$4$}\textcolor{snowblue}{\ding{100}} 
    &&  {$58.33$}  & $64.25/\mathbf{64.79}$& $61.18/\mathbf{62.80}$ & {$\mathbf{61.40}/\mathbf{59.72}$} & {$4$}\textcolor{snowblue}{\ding{100}} 
    && {$30.53$} & {$35.99/\mathbf{37.65}$} & {$35.92/\mathbf{37.23}$} & $\mathbf{37.17}/37.08$    & {$4$}\textcolor{snowblue}{\ding{100}}  \\ 
     
    \midrule

\end{tabular}
}
\end{adjustbox}
\vspace{-0.2cm}
\end{table*}

\vspace{-0.2em}
\begin{itemize}[leftmargin=*]
    \item \textbf{RQ1.} Can Hetero-S boost the performance of prevailing homophilic \& heterophilic GNNs on heterophilic graphs?
    \item \textbf{RQ2.} Does Hetero-S facilitate heterophilic GNNs to extend to more deep network structures?
    \item \textbf{RQ3.} Can Hetero-S genuinely achieve graph sparsity and accelerate computations compared to mainstream graph pruning algorithms \cite{OAR, CGE, chen2021unified, lee2018snip, anonymous2023graph}?
    \item \textbf{RQ4.} How sensitive is Hetero-S to its key components?
\end{itemize}
\vspace{-0.2em}

\noindent To provide answers to these questions, we orchestrate the experiments including \textbf{Main experiments}, \textbf{Depth scalability experiments}, \textbf{Comparative analysis with traditional Snowflake Hypotheses (SnoH)} and \textbf{Efficiency comparison with pruning algorithms} four parts. Detailed descriptions can be found in Appendix \ref{esadd}. Through these experiments, we anticipate drawing clear conclusions regarding the efficacy of Hetero-S.

\vspace{-0.4em}
\subsection{Experiment Setup}

\textbf{Datasets.} We verify Hetero-S across 10 graph benchmarks, including citation networks: Cora, CiteSeer, and PubMed \cite{kipf2016semi}; WebKB networks: Cornell, Texas, and Wisconsin \cite{pei2020geom}; Wikipedia-derived networks: Chameleon and Squirrel \cite{rozemberczki2021multi}; the actor co-occurrence network Actor \cite{pei2020geom}; the heterogenous information network DBLP \cite{wang2019heterogeneous}.  Table \ref{tab:dataset_statistics} in Appendix \ref{data_backbones} offers a comprehensive overview of dataset details. Note that we choose both highly homophily graphs with $\mathcal{H}_{\text{node}} > 0.8$ and heterophilic graphs with $\mathcal{H}_{\text{node}} > 0.1$.

\noindent \textbf{Backbones.} We select three categories of GNN designs, including the \emph{non-local neighbor extension}, \emph{GNN architecture refinement} as stated in Section~\ref{sec:intro}, along with some general GNN backbones. Specifically, for non-local neighbor extension, we choose Mixhop \cite{abu2019mixhop} and GPNN \cite{yang2022graph}. For GNN architecture refinement designs, we opt for backbones like GAT \cite{velivckovic2017graph}, H2GCN \cite{zhu2020beyond}, GCNII \cite{chen2020simple}, Geom-GCN \cite{pei2020geom}, JKNet \cite{xu2018representation}, and MGNN \cite{cui2023mgnn}. Lastly, we choose some general-purpose GNNs, such as GCN \cite{kipf2016semi} and GIN \cite{xu2018powerful}, to further validate the universality of our algorithm.

\vspace{-0.4em}
\subsection{Main Results (RQ1)} \label{RQ1}

\noindent We initially investigate the presence and identifiability of the heterophily snowflake hypothesis (Hetero-S) through the heterophily-aware early stopping (HES) mechanism. We evaluate HES in conjunction with selected GNN backbones across 10 graph benchmarks. Our tests span not only the standard homophilic datasets but also extend to heterophilic graphs. From the Table~\ref{tab:oversmoothing} and Figure~\ref{fig:rq1}, we list the following \textbf{Obs}ervations:

\begin{figure}[h]
  \centering
  \includegraphics[width=1.0\linewidth]{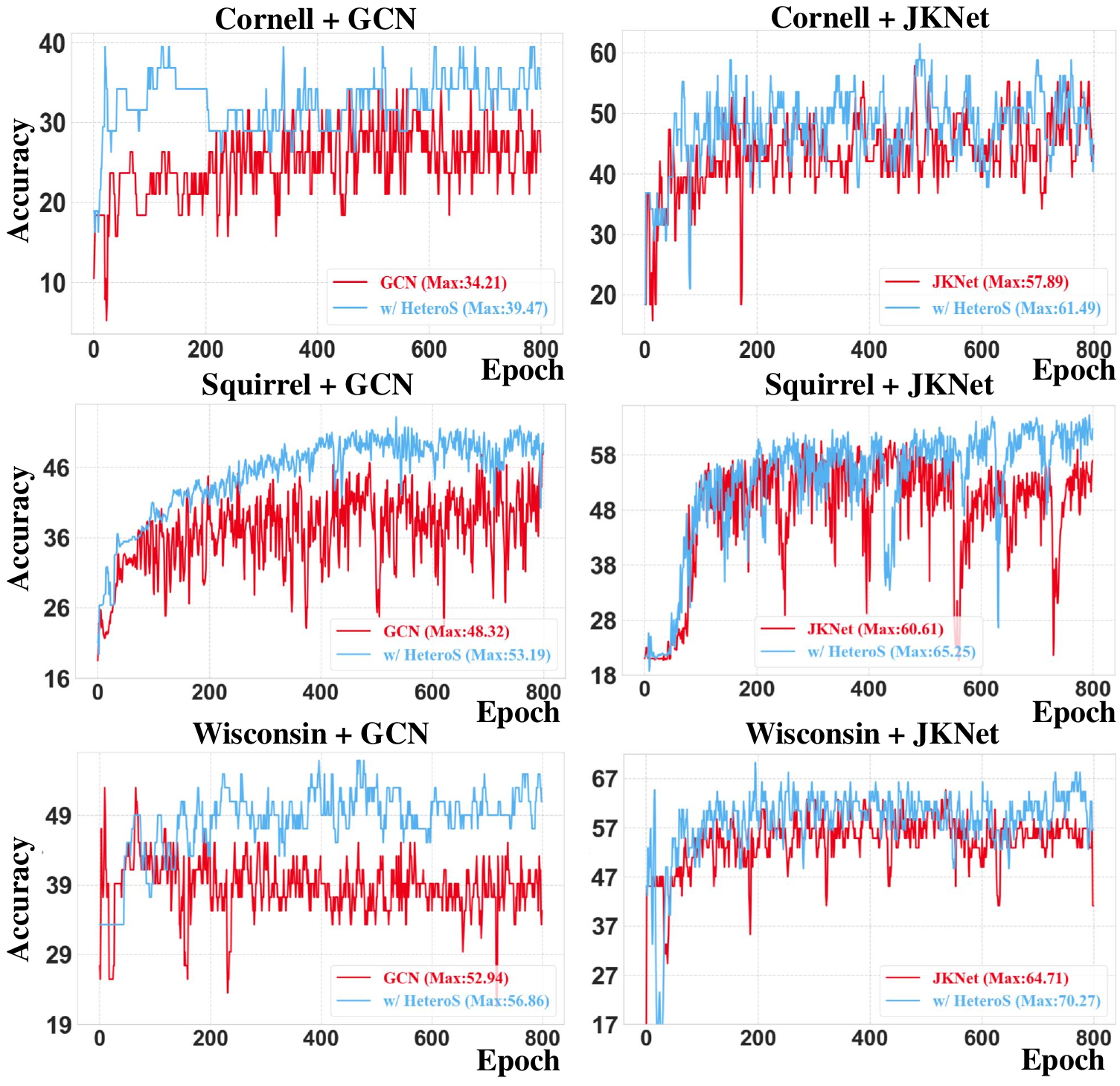}
  \vspace{-0.8cm}
  \caption{The original baselines and +{\textcolor{snowblue}{\ding{100}}} results across Cornell, Squirrel and Wisconsin three benchmarks on 8-layer settings. }
  \label{fig:rq1}
  \vspace{-2.3em}
\end{figure}

\noindent \textbf{Obs.1.} \textbf{The snowflake ({\textcolor{snowblue}{\ding{100}}}) broadly exist under $2\sim8$ layer backbones settings.} As shown in Table \ref{tab:oversmoothing}, upon implementing the HES algorithm, the model consistently achieved performance improvements. For instance, under the MGNN+Cora setup, the model at 8 layers remarkably outperformed the 2-layer baseline by 18.08\%. This phenomenon was also observed across MGNN+Citeseer, JKNet+Texas, H2GCN+Wisconsin, and many others, where performance enhancements ranged from 2.02\%$\sim$14.99\% over the original 2-layer configurations. These findings substantiate the validity of our ``snowflake hypothesis" in heterophily graphs.

\noindent \textbf{Obs.2.} \textbf{HES algorithm showcases the great flexibility to various backbones and consistently presents superior performance.} The introduction of HES consistently results in performance enhancements across nearly all tested models and data combinations. Specifically, with GCN and GIN on homophily graphs like Cora, Citeseer, and PubMed, the models exhibit a $\sim$1\% performance improvement. This trend is even more pronounced in heterophilic graphs where, for example, GCN+Texas/Wisconsin shows an average performance increase of 5.94\% across $2\sim8$ layers. These consistent improvements across various configurations confirm the effectiveness of our proposed HES.

\noindent \textbf{Obs.3.} \textbf{Graph-specific and GNN-specific analyses.} Take a detailed look and analysis of heterophilic graphs combined with tailor-made GNN architectures, we observed that HES contributes significantly to performance gains. For instance, MGNN on 6 datasets with $\mathcal{H}_{node}<0.5$, such as Texas and Squirrel, can yield performance improvements close to 10\%. H2GCN+Texas achieves an increase of approximately $15\%\sim20\%$. Further scrutiny of Figure \ref{fig:rq1} reveals that +{\textcolor{snowblue}{\ding{100}}} feature significantly bolsters the robustness of our training process. Moreover, this enhancement facilitates the discovery of superior dependable subgraphs, particularly within the deeper layers of the network. We showcase more training visualizations in Appendix \ref{app_rq1}.

\vspace{-0.4em}
\subsection{Extend Hetero-S to Deep GNNs (RQ2)} \label{RQ2}
\vspace{-0.2em}
To provide a scalable solution for large and densely connected heterophilic graphs, we extend the HES algorithm to deep GNN contexts. Specifically, we select ResGCN \cite{li2018deeper}, JKNet \cite{xu2018representation}, GCNII \cite{chen2020simple} and SGC \cite{wu2019simplifying} as backbones and conduct tests on 2 homophilic and 2 heterophilic graphs, assessing performance at depths of up to 32 layers. For homophilic graphs, we opt for CS and DBLP~\cite{shchur2019pitfalls}, whose $\mathcal{H}_{\text{node}}$ is 0.81 and 0.82, respectively. For heterophilic graphs, we choose Actor 
 ($\mathcal{H}_{\text{node}}=0.22$) and Chameleon ($\mathcal{H}_{\text{node}}$=0.23),  We list the following observations:

\begin{figure}[!t]
  \centering
  \includegraphics[width=1.0\linewidth]{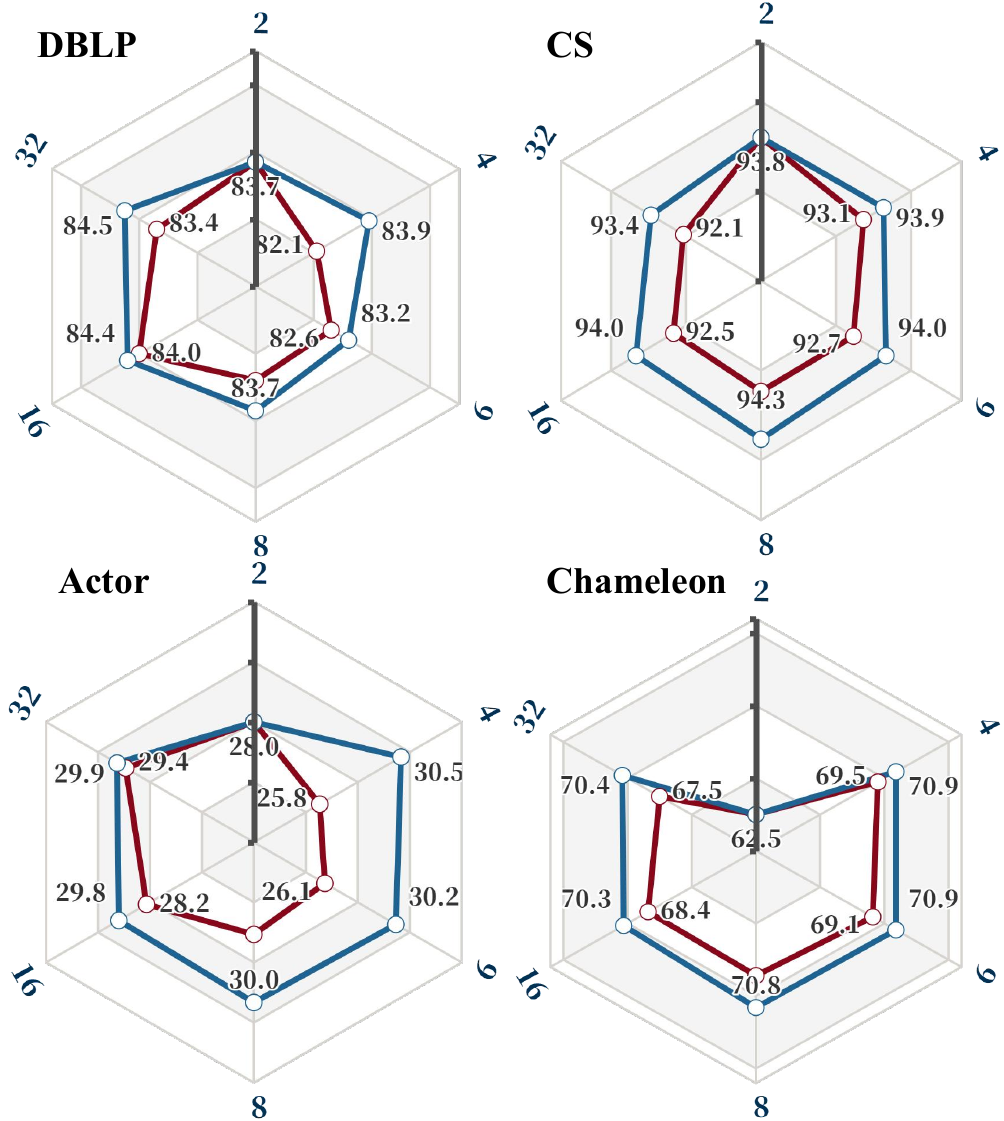}
  \vspace{-0.6cm}
  \caption{The JKNet and +{\textcolor{snowblue}{\ding{100}}} results across CS, DBLP, Actor and Chameleon four benchmarks on 2, 4, 8, 16, 32-layer settings. }
  \vspace{-1.2em}
  \label{fig:rq2}
\end{figure}

\noindent \textbf{Obs.4. Hetero-S consistently boost GNNs at all depths.} As illustrated in Figure \ref{fig:rq2}, the blue line represents the enhanced model performance with the addition of HES, while the red line depicts the original baseline. We observed that incorporating the HES algorithm leads to performance gains in both homophilic and heterophilic graph contexts. Notably, for the \textbf{Chameleon} dataset, the integration of the HES algorithm resulted in a performance increase of nearly 2.9\% against the JKNet baseline.

\noindent \textbf{Obs.5. Hetero-S can assist the ``top-student'' backbones.} An intriguing observation is that JKNet does not exhibit significant performance degradation with the deepening of both homophilic and heterophilic graphs. However, even for the specifically deepened network JKNet, HES still demonstrates exceptional auxiliary performance, further proving the importance of early stopping in receptive fields. We have placed additional experimental results in Appendix \ref{app_rq2}, from which we can draw similar conclusions.

Additionally, the conventional SnoH~\cite{wang2023snowflake} is specially designed for deepening GNNs on homophilic graphs, and we provide further comparisons between Hetero-S and SnoH in Appendix \ref{sec:comp_snoh}.

\vspace{-0.7em}
\subsection{Compare With Pruning Methods (RQ3)} \label{RQ4}
\vspace{-0.3em}

In this section, we compare HES with current SOTA pruning methods, UGS \cite{chen2021unified} and DSpar \cite{anonymous2023graph}. We attempt to understand whether Hetero-S can (1) achieve satisfactory sparsity without compromising performance, and (2) genuinely accelerate GNN computations. As shown in Table ~\ref{tab:sparsity} and Figure \ref{fig:rq4}, we can list the observations:

\begin{figure}[!h]
  \centering
  \includegraphics[width=1.0\linewidth]{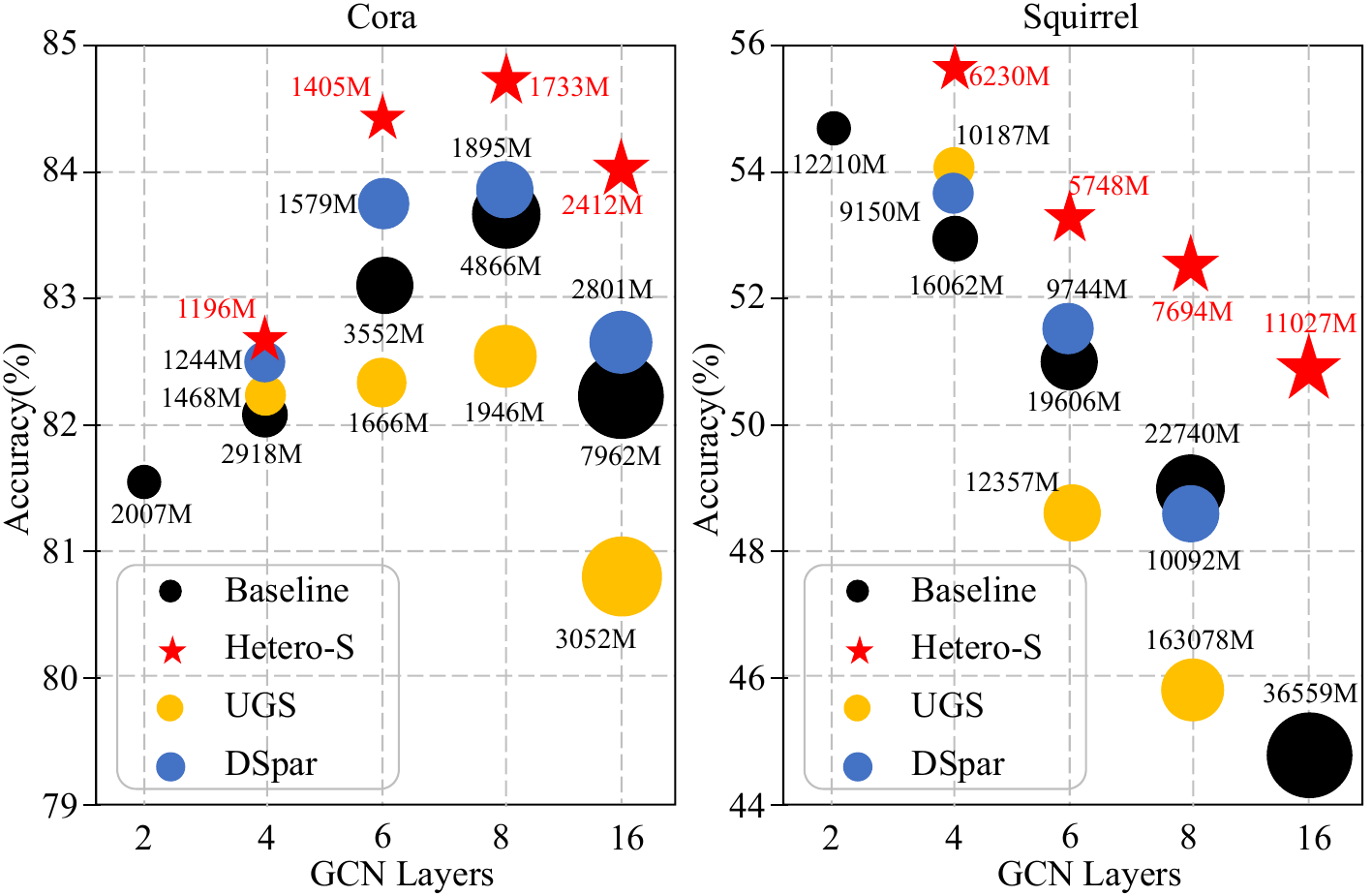}
  \vspace{-0.8cm}
  \caption{Summary of performance (y-axis) at different graph and GNN layers (x-axis) on Cora and Squirrel. The size of markers represents the inference MACs ($=\frac{1}{2}$ FLOPS) of each sparse GCN on the corresponding sparsified graphs. Black circles (\textcolor{black}{$\bullet$}) indicate the baseline. Blue circles (\textcolor{blue}{$\bullet$}) are DSpar. Orange circles (\textcolor{orange}{$\bullet$}) represent the UGS. Red stars (\textcolor{red}{$\star$}) are established by Hetero-S.}
  \vspace{-1.2em}
  \label{fig:rq4}
\end{figure}

\noindent \textbf{0bs.7. Hetero-S consistently achieves the highest sparsity.} As shown in Table~\ref{tab:sparsity}, with the GNN layers deepening, both UGS and DSpar experience a drastic decline in extreme sparsity. In contrast, Hetero-S demonstrates enhanced sparsity capabilities, surpassing UGS and DSpar by 23.96\% and 19.96\% on Squirrel+16-layer GCN.

\noindent \textbf{Obs.8. On both datasets, Hetero-S achieved the smallest inference MACs, nearly only 25\%-45\% of the original baseline.}  Specifically, on the Cora+16-layer GCN, we observe that Hetero-S can achieve comparable or even better performance than UGS and DSpar, with only 8.04\% and 4.89\% MACs, respectively. Furthermore, our algorithm consistently improves performance across various GCN depths, especially in deeper layers. Hetero-S surpassed UGS by approximately 2.3\%$\sim$2.8\% on Cora and was able to reliably train 16 layers on Squirrel, outperforming the baseline by nearly 6.2\%.

\vspace{-0.6em}
\begin{table}[!ht] 
\footnotesize
\setlength{\tabcolsep}{5.0pt}
  \caption{The extreme graph sparsity that HES, UGS, and DSpar are capable of achieving, at which GCN suffers no performance degradation compared with the original baseline.} 
  \vspace{-0.9em}
  \centering
  \begin{tabular}{ccccccccc}
    \toprule
     \textbf{Dataset} & \textbf{Method}  &  \textbf{2}  &  \textbf{4}&   \textbf{6} &  \textbf{8} &  \textbf{16}   \\
    \midrule
       \multirow{3}{*}{Cora}  & UGS
       & $18.55$ & $14.26$ & $14.26$ & 
       $9.75$ & N/A  \\
       
        & DSpar
       & $23.50$ & $21.00$ & $13.00$ & 
       $10.00$ & $7.50$   \\

        & \textbf{HES}
       & \cellcolor{gray!30}$25.18$ & \cellcolor{gray!30}$28.37$ &\cellcolor{gray!30} $30.75$ & \cellcolor{gray!30} $31.98$ & \cellcolor{gray!30}$34.25$   \\
    \midrule
     \multirow{3}{*}{Squirrel}  & UGS
       & $14.26$ & $9.75$ & $5.00$ & 
       $5.00$ & N/A  \\
       
        & DSpar
       & $17.00$ & $12.00$ & $7.00$ & 
       $4.00$ & $4.00$   \\

        & \textbf{HES}
       & \cellcolor{gray!30}$19.11$ & \cellcolor{gray!30}$19.77$ & \cellcolor{gray!30}$20.50$ & \cellcolor{gray!30}$22.89$ & \cellcolor{gray!30}$23.96$   \\

    \bottomrule
  \end{tabular}\label{tab:sparsity}
  \vspace{-1em}
\end{table}

\subsection{Ablation Study (RQ4)} \label{ablation}
\vspace{-0.3em}

Since our performance is contingent on the predictive accuracy of the proxy models, we select multiple proxy models to observe the impact of different proxy predictors on the final prediction results. Concretely, we choose 4-layer GCN and 3-layer GAT, SGC and MLP as proxy predictors to comprehensively verify the model performance. As shown in Table \ref{tab:ablation}, we can make observations:

\begin{table}[!ht] 
\footnotesize
\setlength{\tabcolsep}{8pt}
  \caption{Ablation study on different proxy models.} 
  \vspace{-0.4cm}
  \centering
  \begin{tabular}{ccccccc}
    \toprule
     \textbf{Dataset}   &  \textbf{GCN}  &  \textbf{GAT}&   \textbf{SGC} &  \textbf{MLP}   \\
    \midrule
       Cora   
       & ${82.89_{\pm 0.92}}$ & ${82.11_{\pm 1.24}}$ & ${82.66_{\pm 0.58}}$ & 
       ${82.73_{\pm  0.63}}$   \\
    
      Citeseer  & ${73.34_{\pm  0.82}}$  & ${72.97_{\pm  1.53}}$ & ${73.79_{\pm 1.04}}$ & ${73.40_{\pm 0.93}}$    \\
      
      Texas  
      & ${77.41_{\pm  3.52}}$   
      & ${75.81_{\pm  4.96}}$ 
      & ${77.64_{\pm 2.54}}$ 
      & ${ 78.95_{\pm 2.87}}$ &   \\

      Squirrel  
      & ${55.58_{\pm  1.72}}$
      & ${55.16_{\pm 1.63}}$ & ${56.17_{\pm 1.20}}$ & ${ 56.20_{\pm 1.54}}$ &   \\
      
      Chameleon  
      & ${62.97_{\pm  0.98}}$   
      & ${63.96_{\pm  1.22}}$ 
      & ${63.08_{\pm 0.95}}$
      & ${ 63.39_{\pm 0.89}}$ &   \\
      \midrule
      Avg. Rank & 2.8 & 3.4 & 2.2 & 1.6 &\\
    \bottomrule
  \end{tabular}\label{tab:ablation}
  \vspace{-1.5em}
\end{table}

\noindent \textbf{Obs.9. HES Shows limited sensitivity to proxy model selection.} Across all five datasets, the performance variance of snowflakes obtained using different proxy models ranged narrowly between $0.78\%$ and 1.31\%. Specifically, the overall performance ranking is MLP > SGC > GCN > GAT, so we leverage a 3-layer MLP for the unified experimental setup in all experiments.

\section{Related Work}

Our research primarily focuses on the domain of GNNs and is highly pertinent to two specific areas. Due to space constraints, we have included the comprehensive related work in the appendix \ref{app_related work}.


\noindent \textbf{Graph Pooling \& Clustering} devote to reducing the computational burden of GNNs by applying pruning or compressing methods \cite{chen2018fastgcn, eden2018provable, chen2021unified, eden2018provable, wang2023brave, gao2019graph}, which are highly relevant to our research.  We divide existing techniques into two categories. (1) \emph{Sampling-based methods} aims at selecting the most expressive nodes or edges from the original graph to construct a new subgraph \cite{gao2019graph, lee2019self, ranjan2020asap, zhang2021hierarchical}. Though efficient, the dropping of nodes/edges sometimes results in severe information loss and isolated subgraphs, which may cripple the performance of GNNs \cite{wu2022structural}. (2) \emph{Clustering-based methods} learns how to cluster the whole nodes in the original graph to produces a informative small graph \cite{ying2018hierarchical, wu2022structural, roy2021structure}, which can remedy the aforementioned information loss problem.

\noindent \textbf{Heterophilic GNNs.} Existing heterophilic GNNs primarily fall into two categories: \textit{non-local neighbor extension} and \textbf{GNN architecture refinement} \cite{zheng2022graph}. The former emphasizes expanding the neighborhood scope, achieved via high-order neighbor information mixing \cite{abu2019mixhop, zhu2020beyond, jin2021universal, wang2021tree} and potential neighbor discovery \cite{pei2020geom, liu2021non, yang2022graph}. The latter, delves into enhancing GNNs' expressive power specifically for heterophilic graphs. Strategies include adaptive message aggregation \cite{velivckovic2017graph, gasteiger2018predict}, ego-neighbor separation \cite{zhu2020beyond, suresh2021breaking}, and layer-wise operations \cite{xu2018representation, chen2020simple, chien2020adaptive} to optimize node representation quality. 
It's worth emphasizing that our work shares similarities with that of \cite{wang2023heterophily}, as both approaches utilize proxy models to discern heterogeneity. However, our objective is specifically geared towards pruning the receptive fields that influence aggregation, granting our approach greater versatility. Additionally, our method can better aid in model storage and expedite training.

\vspace{-0.6em}
\section{Conclusion}
In this paper, we first propose the ``one node one receptive field'' concept in heterophilic graph modeling. We further establish the heterophily snowflake hypothesis philosophy for GNNs. To achieve this, we adopt heterophily-aware early stopping to let certain nodes have their own receptive fields. In general, we consistently observe ``snowflakes" across numerous deep architectures. Furthermore, upon testing virtually every type of heterogenous design, we have discovered that our algorithm adeptly integrates with various frameworks, significantly enhancing their performance.

\vspace{-0.6em}
\section{Acknowledgment}

This work is in part supported by the Guangzhou-HKUST(GZ) Joint Funding Program (No. 2024A03J0620).

\bibliographystyle{ACM-Reference-Format}
\bibliography{main}

\appendix

\section{Datasets and backbones descriptions.} \label{data_backbones}

In this section, we provide a detailed description of the datasets and the backbones to enhance the understanding of our experimental design. The statistical characteristics of the datasets are summarized in Table \ref{tab:dataset_statistics}.

\begin{table}[h]
\centering
\caption{The statistics of the datasets.}
\label{tab:dataset_statistics}
\vspace{-0.3cm}
\begin{tabular}{lrrrrr}
\toprule
Dataset & \#Nodes & \#Edges & \#Features & \#Classes & $H_{\text{node}}$ \\
\midrule
CiteSeer & 3,327 & 9,104 & 3,703 & 6 & 0.74 \\
PubMed & 19,717 & 88,648 & 500 & 3 & 0.80 \\
CoraFull & 19,793 & 126,842 & 8,710 & 70 & 0.57 \\
DBLP & 17,716 & 105,734 & 1,639 & 4 & 0.82 \\
CS & 18,333 & 163,788 &  6,805 & 15 & 0.81 \\
\hline
Cornell & 183 & 557 & 1,703 & 5 & 0.13 \\
Texas & 183 & 574 & 1,703 & 5 & 0.09 \\
Wisconsin & 251 & 916 & 1,703 & 5 & 0.19 \\
Chameleon & 2,277 & 62,792 & 2,325 & 5 & 0.23 \\
Squirrel & 5,201 & 396,846 & 2,089 & 5 & 0.22 \\
Actor & 7,600 & 53,411 & 932 & 200 & 0.22 \\

\bottomrule
\end{tabular}
\end{table}

\begin{figure}[h]
\centering
            \includegraphics[scale=0.39]{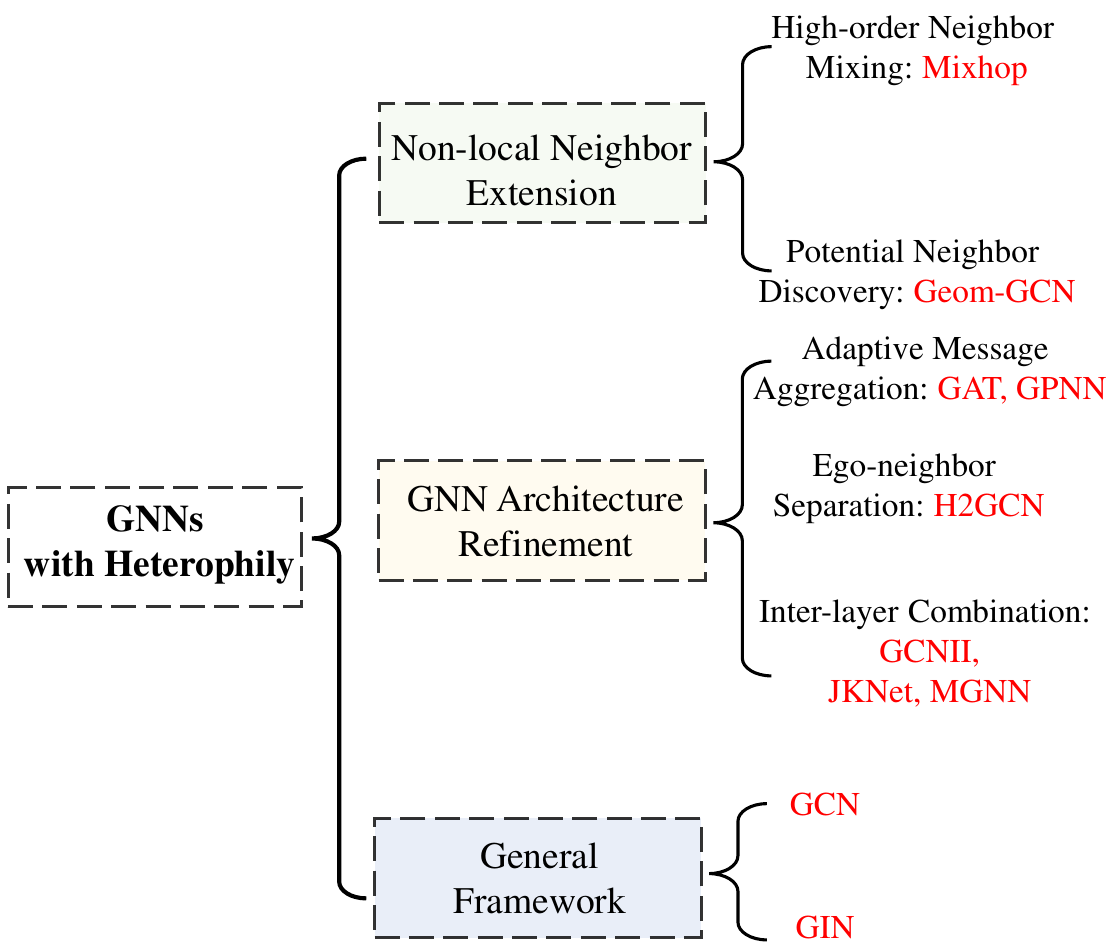}\vspace{-0in}
            \caption{\textbf{Illustration of backbones adopted in our paper.}}\vspace{-0in}
            \label{catego}
\end{figure}

\textbf{Backbone Selection for Validation.} To systematically validate the capability of our ``Heterophily Snowflake Hypothesis," we have selected 10 backbone architectures. We classify our framework into three main categories: Non-local Neighbor Extension, GNN Architecture Refinement, and General Framework. By summarizing these three categories of work, we can more systematically verify the universality of the ``Heterophily Snowflake Hypothesis." Our model categorization is presented as Figure \ref{catego}.

\textbf{High-order Neighbor Mixing.} This approach allows nodes to consider neighbors beyond just the immediate, one-hop neighbors. This seems especially beneficial for heterophilic graphs, allowing them to integrate information from nodes that are more than one edge away.
  
\textbf{Potential Neighbor Discovery.} Instead of just looking at the inherent structure of the graph, this method redefines what a "neighbor" might be. It constructs a potential neighbor set based on some metric function distance in a latent space, rather than just topological closeness.

\textbf{Adaptive Message Aggregation.} Given a set of neighbors, the primary challenge in heterophilic graphs is to effectively aggregate or combine their information. This method seems to alter the aggregation step by weighing the importance of each neighbor differently. The goal is to differentiate information from similar neighbors (of the same class) versus dissimilar neighbors (of different classes).

\textbf{Ego-Neighbor Separation.} The concept of ego-neighbor separation emphasizes differentiating ego node representations from aggregated neighbor nodes for clearer class label distinctions. This approach involves detaching self-loop connections in aggregation and modifying the update function to favor non-mixing operations. 

\textbf{Inter-layer Combination.} Inter-layer combination in GNNs diverges from adaptive message aggregation and ego-neighbor separation methods. Instead of focusing on individual layers, it emphasizes layer-wise operations to enhance GNNs' representation capabilities in heterophily settings. The strategy's foundation is that while shallow GNN layers capture local information, deeper layers grasp broader, global data through repeated neighbor interactions. In a heterophily context, neighbors having similar data might span both immediate vicinity and distant global structures. Thus, integrating representations from every layer optimally leverages diverse neighbor scopes, considering both localized and broad structural characteristics, leading to more robust heterophilic GNNs.

\section{Algorithm Table}

\begin{algorithm}[!ht]
\caption{Algorithm workflow of HES}\label{alg:algo}
\Input{$\mathcal{G}=(\mathbf{A},\mathbf{X})$, GNN model $f_{\mathbf{\Theta}}$, Proxy model $\mathcal{P}_Y$, Epoch number $Q$, GNN layer count $L$}

\For{$\text{iteration}$ i in $\{1,2,\cdots,Q\}$}{
    Forward proxy model and compute $\Tilde{\mathbf{Z}}\leftarrow\mathcal{P}_Y(\mathbf{A},\mathbf{X})$
    
    \tcc{Obtain Homophily mask $\mathbf{S}$}
    
    \For{edge (i,j) in $\mathcal{E}$}{
        $S_{ij}\leftarrow\operatorname{trace}(\Tilde{z}_i \otimes \Tilde{z}_j)$ 
    }
    Compute $k$-hop homophily mask ${\dot{\mathbf{S}}^{(k)}} = {\dot{\mathbf{S}}^k} \in \mathbb{R}^{N \times N}$
    
    Compute the row sum of homophily masks 
    $ set\{ {\mathop \sum_{\text{row}  \setminus \rm{diag}} {{\dot{\mathbf{S}}}^{\left( k \right)}}} \} = \{ \dot{\mathbf{S}}_1^{\left( k \right)},\dot{\mathbf{S}}_2^{\left( k \right)} \ldots \dot{\mathbf{S}}_N^{( k )} \}$

    \For{layer l in $\{1,2,\cdots,L\}$}{
            \tcc{Note that for presentation clarity, we compute embeddings for each node individually here.}
        \For{node $v_i$ in $\mathcal{V}$}{
  
        \uIf{$\dot{S}_i^{(l)} \leq \rho\dot{S}_i^{(1)}$}{
        \tcc{Before receptive early stopping}$\mathbf{h}_i^{(l)}\leftarrow\text{\fontfamily{lmtt}\selectfont \textbf{COMB}}\left( \mathbf{h}_i^{(l-1)}, \text{\fontfamily{lmtt}\selectfont \textbf{AGGR}}\{  \mathbf{h}_j^{(l-1)}: v_j \in \mathcal{N}(v_i) \} \right)$}
                    \Else{\tcc{After receptive early stopping}
$\mathbf{h}_j^{(l)}\leftarrow\text{\fontfamily{lmtt}\selectfont \textbf{COMB}} \left(\mathbf{h}_i^{(l-1)}, \emptyset \right)$
      }
        }

    }
Compute loss function Eq.~\ref{eq:proxy} and Eq.~\ref{eq:opt}

Backward to update GNN $f_{\Theta}$ and proxy model $\mathcal{P}_Y$
    
}

\end{algorithm}

\section{Experimental settings}\label{esadd}
In this section, we report our experimental settings according to the research questions. 

\begin{itemize}[leftmargin=*]
    \item \textbf{Main experiments (RQ1).} In this setup, we integrate Hetero-S into mainstream heterophilic GNNs, focusing on non-local neighbor extensions (2 backbones), GNN architecture refinements (6 backbones) and general designs (2 backbones).
    
    \item \textbf{Depth scalability experiments (RQ2).} We delve into varying depths of GNN architectures. The aim is to determine whether the inclusion of Hetero-S enables these GNNs to maintain or enhance performance as the network goes deeper, avoiding issues like vanishing gradients or over-smoothing.
    
    \item \textbf{Comparative analysis with traditional Snowflake Hypotheses (RQ3).}  Here, we juxtapose Hetero-S with its predecessors, SnoHv1 and SnoHv2, on heterophilic graphs. The experiment is tailored to elucidate if Hetero-S presents a more harmonious alignment with the intricacies of heterophily, potentially leading to better model interpretations and results.

    \item \textbf{Efficiency comparison with pruning algorithms (RQ4).} We compare Hetero-S with current SOTA graph sparsification methods (e.g., UGS \cite{chen2021unified}, SNIP \cite{lee2018snip}, DSpar \cite{liu2023dspar}) with a focus on two key aspects: (1) whether Hetero-S can achieve the desired sparsity without performance compromise, and (2) whether Hetero-S can genuinely accelerate model computations.
\end{itemize}

\section{Additional Results to Answer RQ1.}\label{app_rq1}
In this section, we present additional experiments to answer RQ1. We have included new experimental results for the DBLP dataset and provided further training details for selected datasets.

\begin{table}[h] \small
\setlength{\tabcolsep}{1.5pt} 
\centering
\caption{Quantitative prediction results of DBLP benchmark compared to backbones and (+{\textcolor{snowblue}{\ding{100}}}). The best result is indicated in boldface.}
\label{tab:app_dblp}
\begin{tabular}{l|lcccc}
\toprule
Dataset & Model   & 2-layer\tnote{$\downarrow$} & 4-layer\tnote{$\downarrow$} & 6-layer\tnote{$\uparrow$} & 8-layer \\
\midrule
        & GCN / +{\textcolor{snowblue}{\ding{100}}}    & 83.31/- & 82.97/\textbf{83.33} &  82.09/\textbf{83.24}  & 80.57/\textbf{82.98} \\
        & GIN / +{\textcolor{snowblue}{\ding{100}}} & 80.65/- & 80.27/\textbf{80.33} & 79.39/\textbf{80.35} & 77.87/\textbf{79.90} \\
        
        & GAT / +{\textcolor{snowblue}{\ding{100}}} & 82.51/- & 82.47/\textbf{82.81} & 81.59/\textbf{82.69} & 79.57/\textbf{81.52} \\

        & Mixhop / +{\textcolor{snowblue}{\ding{100}}} & 83.42/- & 83.68/\textbf{84.05} & 82.81/\textbf{83.96} & 81.29/\textbf{83.70} \\
DBLP& Geom-GCN/ +{\textcolor{snowblue}{\ding{100}}} & 82.46/-  & 28.37/ \textbf{44.84} & 26.38/ \textbf{59.36}  & 19.66/\textbf{59.78}   \\
        
        & H2GCN / +{\textcolor{snowblue}{\ding{100}}}   & 83.96/- & 83.92/\textbf{84.95} & 83.34/\textbf{83.97} & 83.22/\textbf{83.68} \\
        
        & GCNII / +{\textcolor{snowblue}{\ding{100}}}    & 82.07/- & 83.16/\textbf{83.89} &  83.17/\textbf{83.66} & 83.36/\textbf{83.87} \\
        
        & APPNP / +{\textcolor{snowblue}{\ding{100}}} & 83.39/- & 83.05/\textbf{84.02} & 82.17/\textbf{83.14} & 80.65/\textbf{81.62} \\

        & JKNet / +{\textcolor{snowblue}{\ding{100}}} & 81.94/- & 82.91/\textbf{83.24} & 82.92/\textbf{83.41} & 83.11/\textbf{83.62} \\
        
        & MGNN / +{\textcolor{snowblue}{\ding{100}}} & 83.84/- & 83.50/\textbf{84.29} & 82.62/\textbf{83.98} & 81.10/\textbf{81.56} \\
\bottomrule
\end{tabular}
\end{table}

\begin{figure*}[h]
\setlength{\abovecaptionskip}{6pt}
\centering
\includegraphics[width=0.95\textwidth]{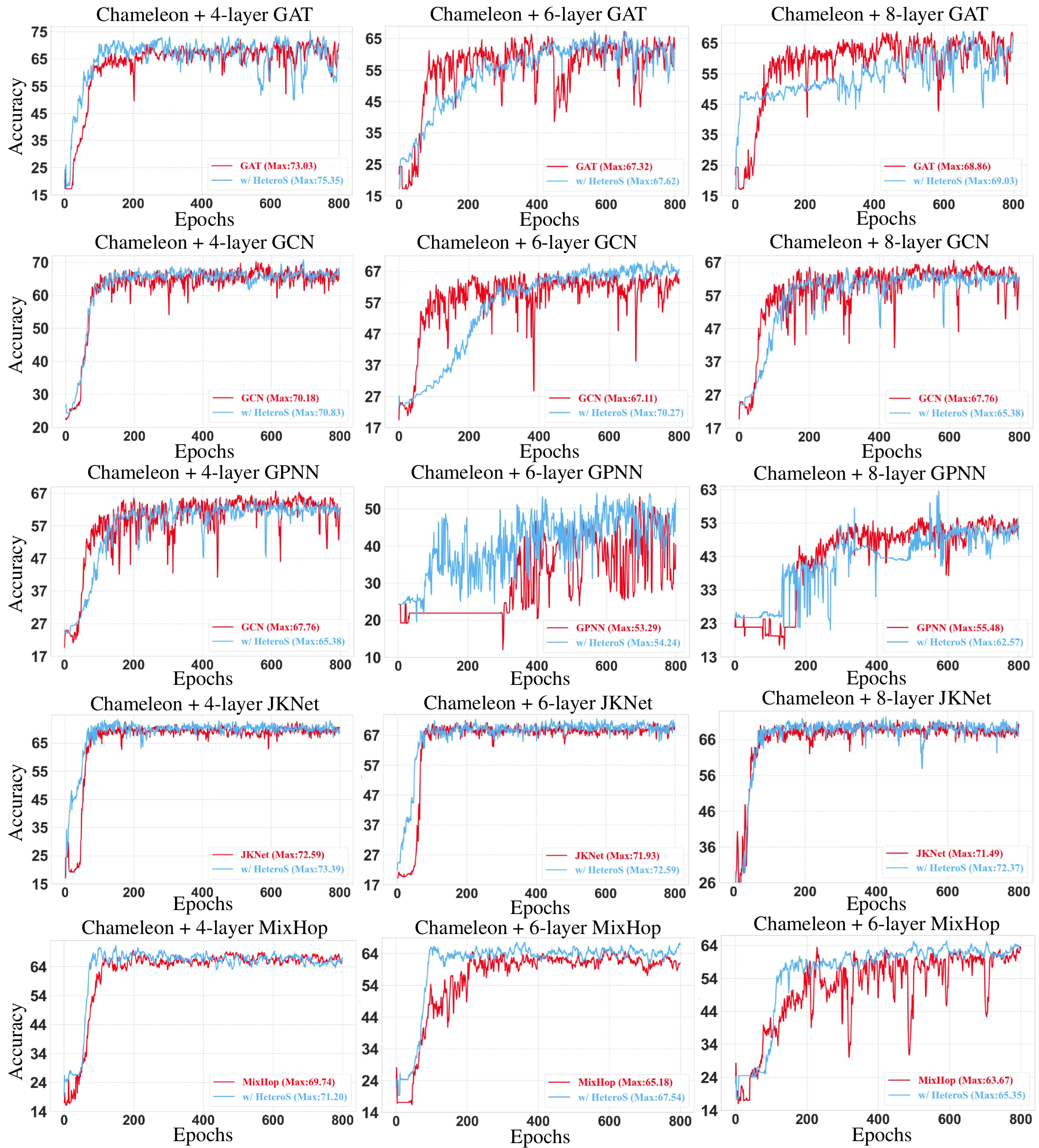}
\caption{The performance of backbones and the results after adding Hetero-S (+{\textcolor{snowblue}{\ding{100}}}).} \label{fig:main_app_1}
\end{figure*}

\begin{figure*}[h]
\setlength{\abovecaptionskip}{6pt}
\centering
\includegraphics[width=0.95\textwidth]{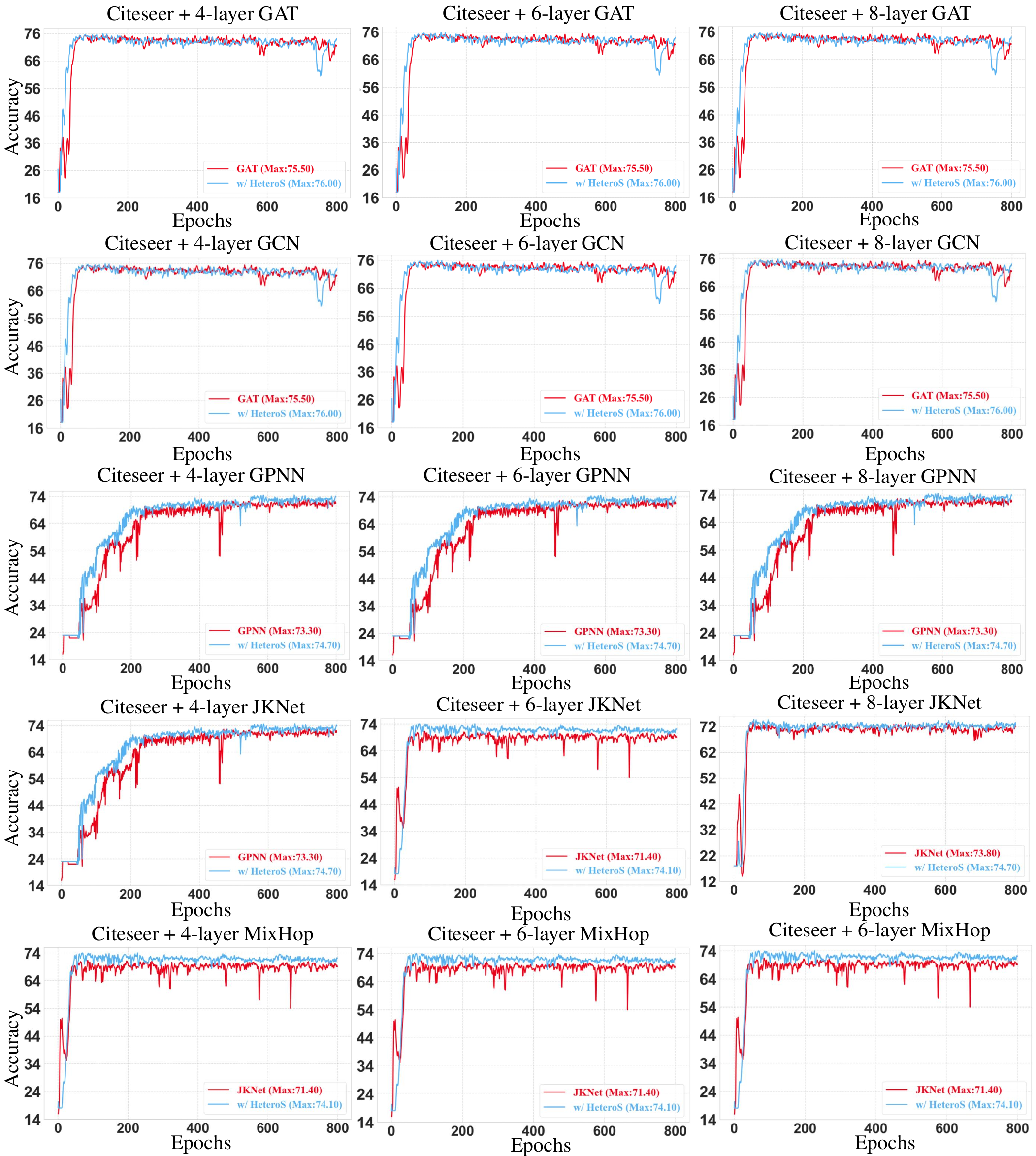}
\caption{The performance of backbones and the results after adding Hetero-S (+{\textcolor{snowblue}{\ding{100}}}).} \label{fig:main_app_2}
\end{figure*}

\begin{figure*}[h]
\setlength{\abovecaptionskip}{6pt}
\centering
\includegraphics[width=0.95\textwidth]{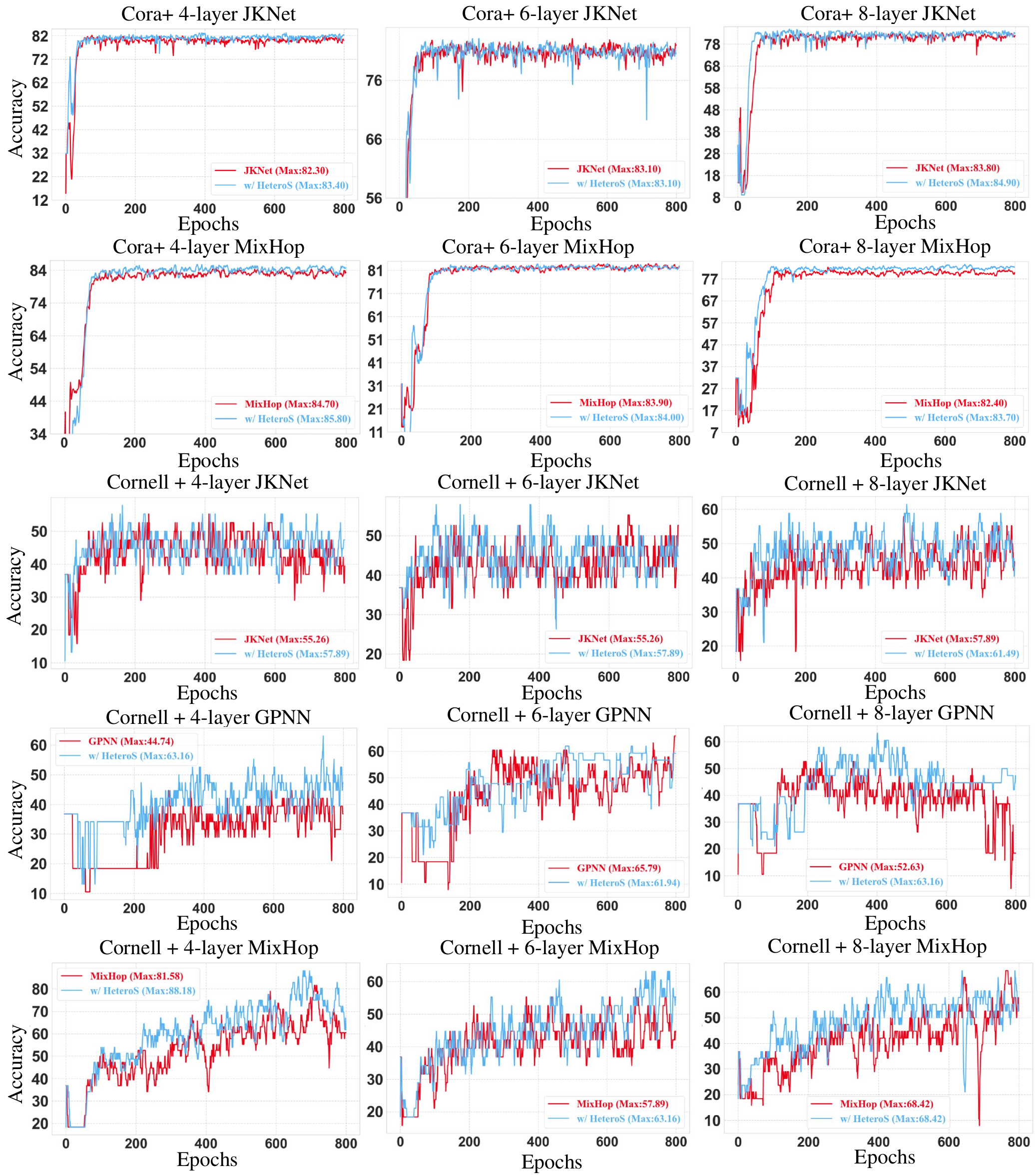}
\caption{The performance of backbones and the results after adding Hetero-S (+{\textcolor{snowblue}{\ding{100}}}).} \label{fig:main_app_3}
\end{figure*}

\begin{figure*}[h]
\setlength{\abovecaptionskip}{6pt}
\centering
\includegraphics[width=0.95\textwidth]{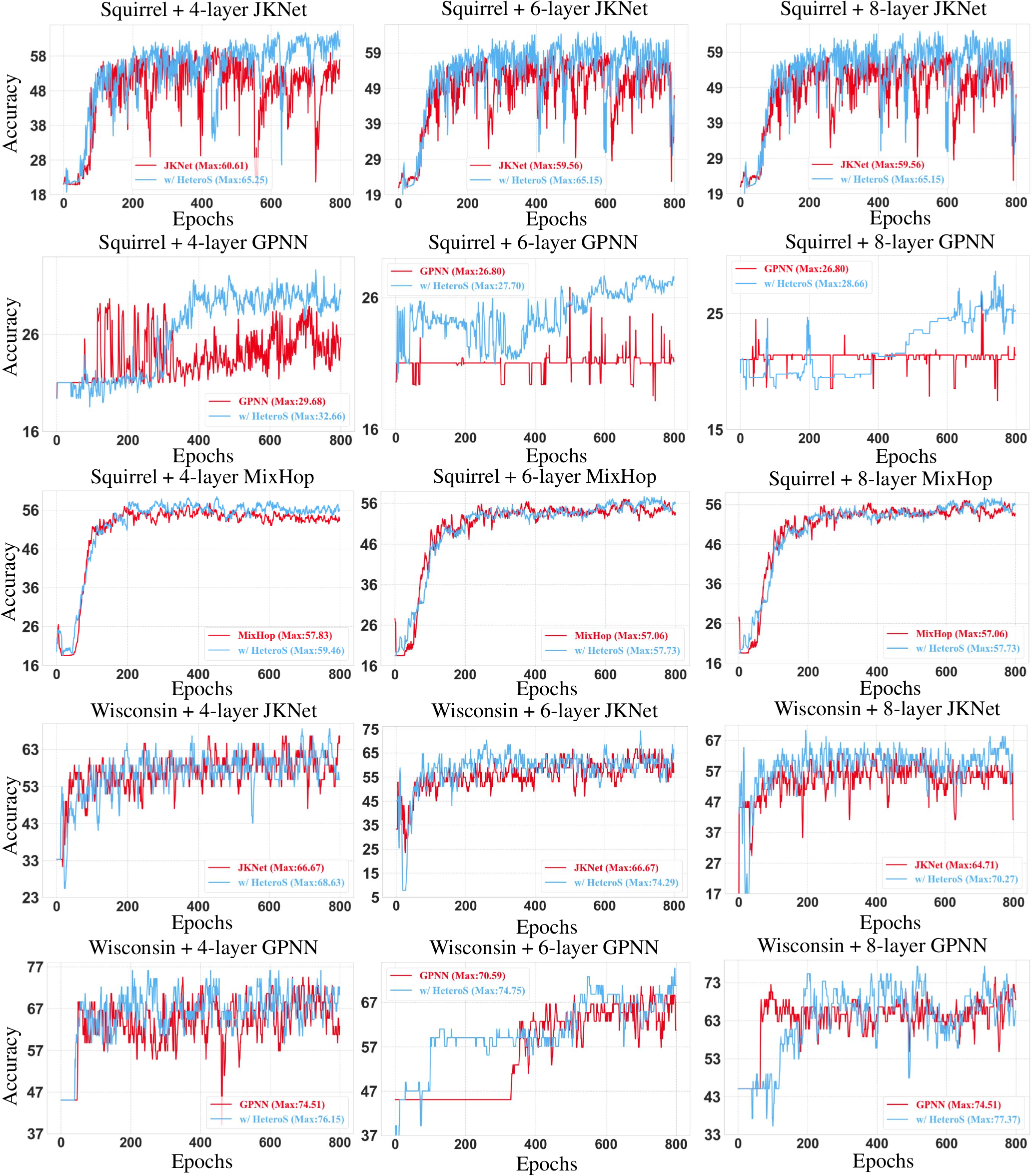}
\caption{The performance of backbones and the results after adding Hetero-S (+{\textcolor{snowblue}{\ding{100}}}).} \label{fig:main_app_4}
\end{figure*}

\noindent As shown in Table \ref{tab:app_dblp}, we list observations (1) Across most backbones, the incorporation of Hetero-S generally improved the prediction results, as seen by the higher scores in the columns with the +{\textcolor{snowblue}{\ding{100}}}. This showcases the potential benefits of integrating Hetero-S into these architectures. (2) Backbones exhibit varied performance across different layer depths. Notably, certain architectures, such as Geom-GCN, experience significant fluctuations in scores as the layers increase. This suggests that while increased depth might amplify the capabilities of some models, it can be counterproductive for others, particularly for GCNII and JKNet. Intriguingly, we observe that upon incorporating Hetero-S, the backbone's performance can even revert to optimal levels exhibited by the original few layers, including the 2-layer structure. This strongly validates the heterophily snowflake hypothesis we propose.

We also present the performance curves for tests incorporating Hetero-S with the original backbone. As shown in Table \ref{fig:main_app_1}, we can observe the test performance curves for traditional backbones and the enhancement achieved by integrating our Hetero-S framework.  Each graph plots the accuracy over the number of epochs, divided into two scenarios: one for the original backbone (in red) and one for the backbone with Hetero-S (in blue). 

\textbf{For the Chameleon dataset}, when comparing the 4-layer, 6-layer, and 8-layer GAT, GCN, GPNN, and JKNet, we can see a consistent pattern. The addition of Hetero-S generally leads to an improvement in maximum test accuracy across all backbone architectures and depths. This is particularly evident in scenarios where the original backbone has a more volatile or lower accuracy trajectory. For instance, in the 4-layer GAT, the peak accuracy improves from 73.53\% to 75.45\% with the inclusion of Hetero-S. Similar improvements are observed in 6-layer and 8-layer configurations. The 6-layer GAT sees an increase from 73.67\% to 76.22\%, and the 8-layer from 85.60\% to 86.93\%. The GCN models also show enhancement with Hetero-S, albeit with a more significant impact on the 4-layer and 6-layer models than on the 8-layer variant. Meanwhile, the GPNN models reflect a notable benefit from Hetero-S at all depths, with the 4-layer model showing an increase from 67.76\% to 68.36\%, the 6-layer from 55.29\% to 58.24\%, and the 8-layer from 54.58\% to 57.57\%. JKNet architectures display similar trends, with the 4-layer model's accuracy increasing from 72.59\% to 73.49\%, the 6-layer from 71.93\% to 74.59\%, and the 8-layer from 74.19\% to 74.37\%. These results suggest that Hetero-S provides a consistent and noteworthy enhancement to the stability and accuracy of various graph neural network architectures across different model complexities. We also showcase more results in Figure \ref{fig:main_app_2} $\sim$ \ref{fig:main_app_4}, from which we can draw similar conclusions from these results. 

\section{Additional Results to Snswer RQ2.}\label{app_rq2}

\begin{figure*}[t]
  \centering
  \includegraphics[width=0.98\linewidth]{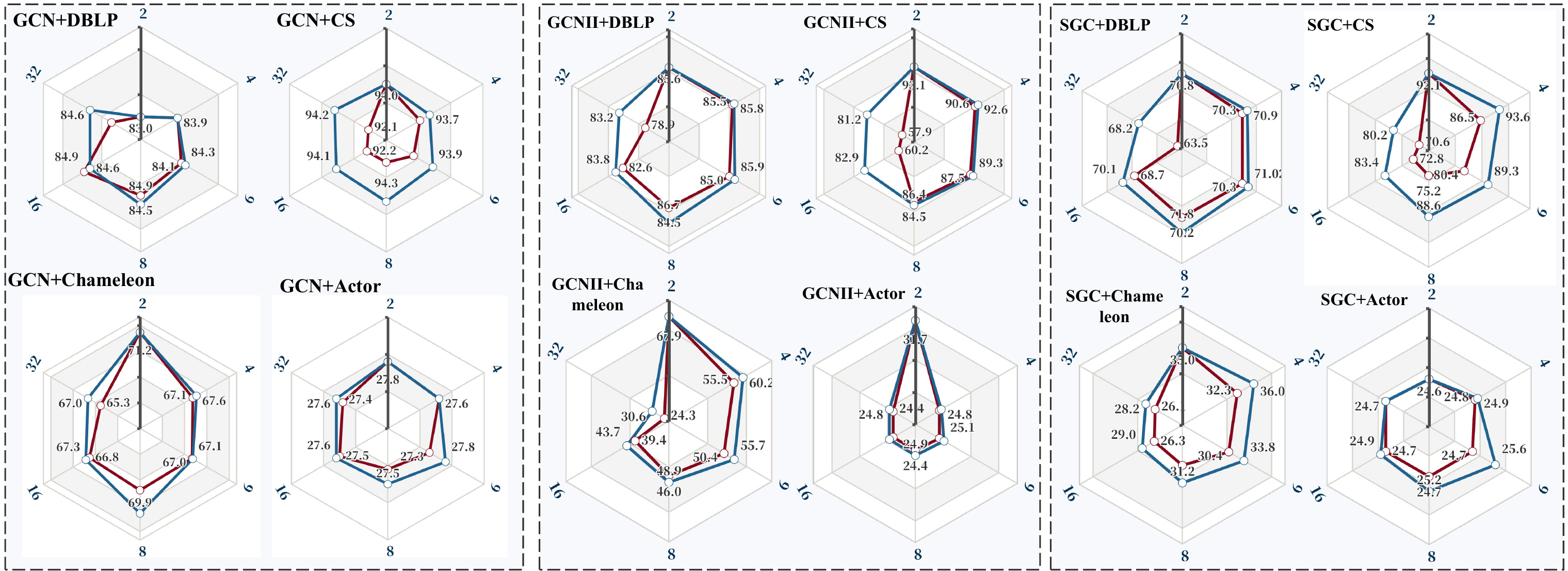}
  \caption{The GCN, GCNII, SGC and +{\textcolor{snowblue}{\ding{100}}} results across CS, DBLP, Actor and Chameleon four benchmarks on 2, 4, 8, 16, 32-layer settings. }
  \label{fig:rq2appen}
  \vspace{-7pt}
\end{figure*}

In this section, we showcase the additional experimental results on DBLP, CS, Chameleon and Actor four graph bechmark. As shown in Figure \ref{fig:rq2appen}, Incorporating the HES algorithm consistently enhances performance across GCN, GCNII, and SGC models. For example, the GCN+DBLP model reveals a significant performance peak at a depth of 8 layers, indicating an optimal balance between model complexity and learning capability. The GCN+CS model consistently performs well, particularly at lower layer counts, suggesting that the HES algorithm captures the essential representational features efficiently, even without deeper network architectures. Moreover, the GCNII variants benefit more markedly from the integration with the HES algorithm, especially evident in the Actor and Chameleon benchmarks. The enhanced performance indicates that the HES algorithm's approach to leveraging heterogeneity in data aligns exceptionally well with the sophisticated architectures of GCNII.

Furthermore, the SGC models also show improvements in performance, indicating the HES algorithm's robustness across different levels of model complexity. This reaffirms its utility as a versatile enhancer of graph network effectiveness. Notably, the performance boosts are not solely dependent on depth but also show a trend of incremental gains with increasing complexity, up to a point where performance begins to plateau or slightly decline, highlighting the HES algorithm's subtle influence on model.

\section{Performance Comparison With Conventional SnoH}\label{sec:comp_snoh}


In this section, our focus is on a systematic examination of the performance merits and limitations of HES, in contrast to conventional ``snowflake'' settings. To facilitate this analysis, we choose six distinct heterophilic graph benchmarks as foundational models for GNNs, specifically: Texas, Wisconsin, Cornell, Squirrel, Chameleon, and Actor. Notably, the depth of heterophilic GNNs is typically observed to not exceed five layers, as indicated by \cite{yan2022two}. Hence, within the framework of our heterophilic data approach, we deliberately confine the depth of our network to 6 and 8 layers. This strategic limitation enables us to conduct a thorough and precise comparative analysis of the performances of SnoHv1, SnoHv2, and Hetero-S. In order to conduct a comprehensive comparison, we employ both a homophilic (JkNet) and a heterophilic (Mixhop) GNN as the backbones for our experimental analysis.

\begin{table}[h] 
\footnotesize
\setlength{\tabcolsep}{1.8pt}
  \caption{Comparison results among different ``snowflake'' methods across 6 and 8-layers under Mixhop backbone setting.} 
  \centering
  \scalebox{0.92}{
  \begin{tabular}{cccccccc}
    \toprule
     \multirow{2}{*}{\makecell{\textbf{Benchmark}\\(JKNet)}}   &   \multirow{2}{*}{\textbf{$H_{node}$}}   &  \multicolumn{3}{c}{\textbf{6-layer}}  &  \multicolumn{3}{c}{\textbf{8-layer}}  \\
     
     \cmidrule(lr){3-5} \cmidrule{6-8} 
     
     & & \textbf{SnoHv1} &  \textbf{SnoHv2}  &  \textbf{Hetero-S} & \textbf{SnoHv1} &  \textbf{SnoHv2}  &  \textbf{Hetero-S}  \\
    \midrule
    Texas   & 0.11  & ${82.78_{\pm 4.17}}$ & ${83.26_{\pm  3.77}}$ & \cellcolor{gray!25}${84.30_{\pm 3.89}}$  & ${86.25_{\pm  4.33}}$ &${86.50_{\pm 4.29}}$ & \cellcolor{gray!25}${89.88_{\pm 4.03}}$ \\
    
    Wisconsin  & 0.21    & ${60.72_{\pm  5.88}}$  & ${59.07_{\pm  5.12}}$ & \cellcolor{gray!25}${64.99_{\pm  5.05}}$ & ${60.78_{\pm  3.93}}$ & ${59.21_{\pm  4.20}}$ & \cellcolor{gray!25}${63.82_{\pm  3.51}}$ \\
      
    Cornell & 0.22   & ${48.22_{\pm  3.87}}$  & ${48.70_{\pm  6.85}}$ & \cellcolor{gray!25}${50.07_{\pm  3.25}}$ & ${47.15_{\pm  4.28}}$ & ${47.92_{\pm  5.04}}$ & \cellcolor{gray!25}${48.65_{\pm  3.98}}$   \\

    Squirrel  & 0.22   & ${59.64_{\pm  2.15}}$  & ${58.77_{\pm  1.77}}$ & \cellcolor{gray!25}${61.22_{\pm  1.33}}$ & ${59.88_{\pm  1.82}}$ & ${59.04_{\pm  1.54}}$ & \cellcolor{gray!25}${61.48_{\pm  1.56}}$  \\

    Chameleon  & 0.23   & ${68.93_{\pm  1.78}}$  & ${68.56_{\pm  2.05}}$ & \cellcolor{gray!25}${69.44_{\pm  1.62}}$ & ${69.42_{\pm  2.15}}$ & ${68.77_{\pm  2.24}}$ & \cellcolor{gray!25}${70.93_{\pm  1.66}}$    \\

    Actor  & 0.22   & ${28.76_{\pm  2.01}}$  & ${28.65_{\pm  1.94}}$ & \cellcolor{gray!25}${30.04_{\pm  1.78}}$ & ${28.55_{\pm  2.86}}$ & ${27.93_{\pm  2.30}}$ & \cellcolor{gray!25}${29.65_{\pm  1.96}}$    \\

    \midrule
    \multirow{2}{*}{\makecell{\textbf{Benchmark}\\(MixHop)}}   &   \multirow{2}{*}{\textbf{$H_{node}$}}   &  \multicolumn{3}{c}{\textbf{6-layer}}  &  \multicolumn{3}{c}{\textbf{8-layer}}  \\
     
     \cmidrule(lr){3-5} \cmidrule{6-8} 
     
     & & \textbf{SnoHv1} &  \textbf{SnoHv2}  &  \textbf{Hetero-S} & \textbf{SnoHv1} &  \textbf{SnoHv2}  &  \textbf{Hetero-S}  \\
    \midrule
    Texas   & 0.11  & ${86.96_{\pm  2.88}}$ & ${87.66_{\pm  1.79}}$ & \cellcolor{gray!25}${89.68_{\pm 1.43}}$  & ${73.33_{\pm  3.80}}$ &${74.19_{\pm  4.65}}$ & \cellcolor{gray!25}${76.32_{\pm 3.17}}$ \\
    
    Wisconsin  & 0.21    & ${76.92_{\pm  4.16}}$  & ${78.45_{\pm  3.99}}$ & \cellcolor{gray!25}${82.93_{\pm  3.27}}$ & ${75.48_{\pm  3.88}}$ & ${76.66_{\pm  3.14}}$ & \cellcolor{gray!25}${80.28_{\pm  3.36}}$ \\
      
    Cornell & 0.22  & ${55.88_{\pm  3.96}}$  & ${57.40_{\pm  3.79}}$ & \cellcolor{gray!25}${64.32_{\pm  3.56}}$ & ${41.31_{\pm  3.68}}$ & ${49.63_{\pm  3.74}}$ & \cellcolor{gray!25}${65.70_{\pm  3.06}}$  \\

    Squirrel  & 0.22  & ${53.12_{\pm  0.87}}$  & ${53.70_{\pm  1.29}}$ & \cellcolor{gray!25}${53.67_{\pm  1.33}}$ & ${53.09_{\pm  1.26}}$ & ${52.76_{\pm  0.73}}$ & \cellcolor{gray!25}${52.87_{\pm  0.96}}$ \\

    Chameleon  & 0.23  & ${63.16_{\pm  2.23}}$  & ${64.78_{\pm  2.77}}$ & \cellcolor{gray!25}${66.04_{\pm  2.50}}$ & ${62.79_{\pm  2.18}}$ & ${63.46_{\pm  2.94}}$ & \cellcolor{gray!25}${65.89_{\pm  2.66}}$    \\

    Actor  & 0.22  & ${35.87_{\pm  0.94}}$  & ${36.02_{\pm  1.08}}$ & \cellcolor{gray!25}${36.68_{\pm  0.66}}$ & ${35.79_{\pm  0.98}}$ & ${35.11_{\pm  0.91}}$ & \cellcolor{gray!25}${35.28_{\pm  0.79}}$   \\
    \bottomrule
  \end{tabular}\label{tab:trans}
  }
  \vspace{-1em}
\end{table}

\noindent \textbf{Obs.6.} \textbf{Hetero-S consistently outperforms the other methods (SnoHv1 and SnoHv2) in most benchmarks for both 6-layer and 8-layer configurations.} Particularly in the 8-layer setting under the MixHop backbone, Hetero-S demonstrates a significant advantage, suggesting that our model scales well with depth and benefits from the MixHop architecture. For instance, in the Texas benchmark, Hetero-S shows a remarkable improvement from $76.32\pm3.17$ to $89.88\pm4.03$, indicating its robustness in deeper network structures.


\section{Parameter Sensitivity Analysis}\label{sec:hyper}

In this section, we detail how we determine the filtering threshold $\rho$ in all experiments and how $\rho$ influences the performance of HES. In practice, we want to avoid both excessively large $\rho$, as it could lead to premature removal of too large a receptive field during early stopping, resulting in suboptimal model performance, and excessively small $\rho$, as they could cause central nodes to absorb too much heterophilic information from multi-hop neighbors. Therefore, we search the most suitable $\rho$ in a limited range \{1e-2, 1e-4, 1e-6, 1e-8, 1e-16\} for all experiments. To further demonstrate how sensitive our method is to $\rho$, we test the performance of HES on Squirrel/Chameleon with different filtering threshold settings. As shown in Table~\ref{tab:rho}, we can observe (1) the optimal performance of HES requires an appropriate choice of $\rho$. Generally, as $\rho$ increases, HES's performance often initially improves (corresponding to increased removal of receptive fields), then declines (corresponding to excessive removal of receptive fields); (2) Overall, HES is relatively insensitive to the choice of $\rho$. For instance, on GCN, HES performance varies by no more than 1.04\% and 1.34\%.

\begin{table}[!ht] 
\footnotesize
\setlength{\tabcolsep}{4pt}
  \caption{Parameter sensitivity on filtering threshold $\rho$. We report the performance with HES under 4-layer settings with different $\rho$.} 
  \centering
  \begin{tabular}{cccccc}
    \toprule
     \textbf{Squirrel}   &  \textbf{1e-2}  &  \textbf{1e-4}&   \textbf{1e-6} &  \textbf{1e-8}   &  \textbf{1e-16}   \\
    \midrule
      GCN+{\textcolor{snowblue}{\ding{100}}}  
       & $55.18$ & $56.20$ & $55.52$ & 
       $55.78$ & $54.74$  \\
    
      MixHop+{\textcolor{snowblue}{\ding{100}}} & $55.42$  & $56.88$ & $55.46$ & $55.20$ & $54.82$   \\
      
      JKNet+{\textcolor{snowblue}{\ding{100}}}  
      & $59.77$   
      & $60.11$ 
      & $60.08$ 
      & $60.02$ 
      & $58.55$  \\

      MGNN+{\textcolor{snowblue}{\ding{100}}}  
      & $45.65$
      & $46.81$ & $47.85$ & $47.23$ 
      & $ 47.56$   \\
    \midrule
         \textbf{Chameleon}   &  \textbf{1e-2}  &  \textbf{1e-4}&   \textbf{1e-6} &  \textbf{1e-8}   &  \textbf{1e-16} \\
    \midrule
      GCN+{\textcolor{snowblue}{\ding{100}}}  
       & $62.25$ & $63.11$ & $63.38$ &
       $63.23$  &  $62.34$\\
      
      MixHop+{\textcolor{snowblue}{\ding{100}}} 
      & $67.74$  
      & $69.04$ 
      & $68.78$ 
      & $69.08$ 
      & $67.22$ \\
      
      JKNet+{\textcolor{snowblue}{\ding{100}}}  
      & $70.17$   
      & $70.38$ 
      & $72.37$ 
      & $71.69$ 
      & $70.55$   \\

      MGNN+{\textcolor{snowblue}{\ding{100}}}  
      & $60.19$
      & $62.09$ & $61.76$ & $61.92$ & $60.83$  \\
    \bottomrule
  \end{tabular}\label{tab:rho}
\end{table}

\section{Case Study}\label{case}

\begin{figure*}[h]
  \centering
  \includegraphics[width=0.90\linewidth]{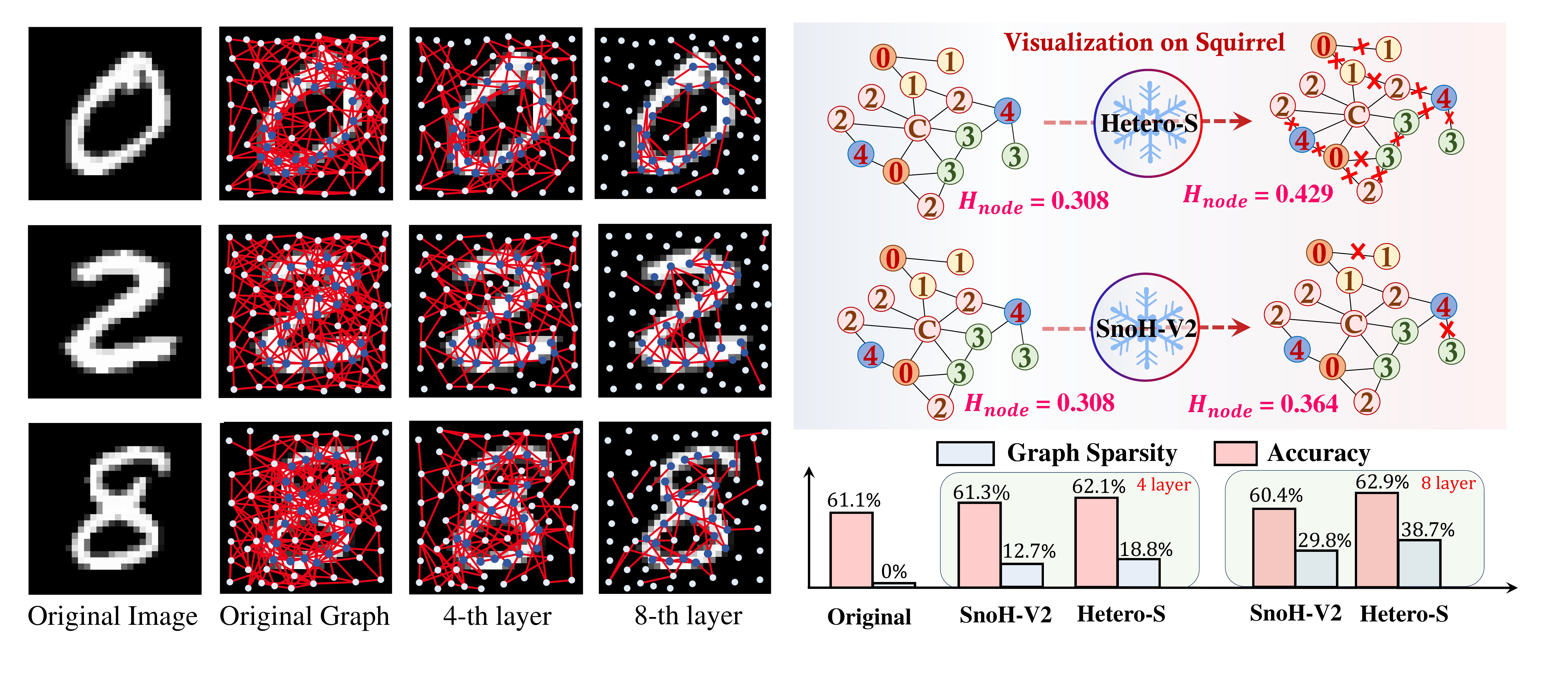}
  \caption{(\textbf{\textit{Left.}}) Visualization of the subgraphs extracted by applying HES to an 8-layer GCN with MNIST. Original images and graphs are displayed on the first and second columns. Visualization of the subgraphs extracted by applying HES to a 4-layer GCN with Texas. The central (\textbf{\textit{Right.}}) Prediction results for the central node C (The label is 2) using different algorithms.}
  \label{fig:case_study}
  \vspace{-7pt}
\end{figure*}

In this section, we endeavor to investigate the efficacy of Hetero-S from a micro-perspective through the analysis of two case studies. We choose the superpixel graphs of MNIST \cite{knyazev2019understanding} and Squirrel as benchmarks to observe the visualized outcomes. As depicted in Figure \ref{fig:case_study}, the following observations can be made:


\begin{itemize}[leftmargin=*]
    \item \textbf{\textit{Left.}} On the MNIST dataset, we observe that as the depth of the GNN increases, the edges in the black regions are progressively pruned. At the deepest layer, the adjacency matrix aligns precisely with the white regions, capturing the edge information of the prediction areas, thereby significantly aiding the model's predictive capability. This validates that the HES algorithm adeptly captures the most critical information for prediction and eliminates redundant information.

    \item \textbf{\textit{Right.}} In our analysis of the Squirrel, we find that the conventional SnoHv2, after pruning, increases its homophily ratio from 0.308 to 0.364. Conversely, Hetero-S enhances this ratio to 0.429, proving that the \textit{Hetero-S scheme improves the model's likelihood of aggregating similar labels, thereby boosting the effectiveness of information aggregation}. Hetero-S achieves a dual victory of greater graph sparsity and accuracy compared to SnoH-v2, further attesting to the superior capabilities of HES over SnoH-v2 in heterophilic graph scenarios. For instance, HES consistently surpasses SnoH-v2 by substantial performance margins across Squirrel dataset ($0.8\%\sim2.5\% \uparrow$ on accuracy and $6.1\%\sim8.9\% \uparrow$ on graph sparsity).
\end{itemize}

\section{Related Work} \label{app_related work}

\textbf{Graph Neural Networks (GNNs).} GNNs have emerged as a prominent subfield in machine learning, specifically tailored to manage and analyze graph-structured data \cite{scarselli2008graph, wu2020comprehensive}. In general, GNNs owe their efficacy to a distinct ``message-passing'' mechanism, which seamlessly integrates topological structures with node characteristics to yield richer graph representations. This process is best described by the mathematical expression $H^{(k)} = M(A, H^{(k-1)}, \theta^{(k)})$. In this equation, $H^{(k)}$ stands for the node embedding after $k$ iterations of GNN aggregation. Meanwhile, $M$ represents the message propagation function, and $\theta^{(k)}$ denotes the trainable parameters at a given layer \cite{xu2018powerful, fan2019graph, liu2020towards}. The escalating enthusiasm surrounding GNNs has catalyzed the development of a myriad of propagation techniques \cite{zhu2019aligraph, jiang2022graph, ying2019gnnexplainer} and model variants \cite{wang2022searching, yu2020representative, thekumparampil2018attention, you2019position}, which have significantly broadened the arsenal of tools available for graph-oriented learning and exploration.

\noindent \textbf{Graph Pooling \& Clustering.} Graph pooling and clustering devote to reducing the computational burden of GNNs by applying pruning or compressing methods \cite{chen2018fastgcn, eden2018provable, chen2021unified, eden2018provable, chen2021unified, gao2019graph}, which are highly relevant to our research.  We divide existing techniques into two categories. (1) \emph{Sampling-based methods} aims at selecting the most expressive nodes or edges from the original graph to construct a new subgraph \cite{gao2019graph, lee2019self, ranjan2020asap, zhang2021hierarchical}. Though efficient, the dropping of nodes/edges sometimes results in severe information loss and isolated subgraphs, which may cripple the performance of GNNs \cite{wu2022structural}. (2) \emph{Clustering-based methods} learns how to cluster the whole nodes in the original graph, and produces a informative graph where the clusters are node sets \cite{ying2018hierarchical, wu2022structural, roy2021structure}, which can remedy the aforementioned information loss problem.

\noindent \textbf{Heterophilic GNNs.} Existing heterophilic GNNs primarily fall into two categories: \textit{non-local neighbor extension} and \textbf{GNN architecture refinement} \cite{zheng2022graph}. The former emphasizes expanding the neighborhood scope, achieved via high-order neighbor information mixing \cite{abu2019mixhop, zhu2020beyond, jin2021universal, wang2021tree} and potential neighbor discovery \cite{pei2020geom, liu2021non, yang2022graph}. The latter, delves into enhancing GNNs' expressive power specifically for heterophilic graphs. Strategies include adaptive message aggregation \cite{velivckovic2017graph, gasteiger2018predict}, ego-neighbor separation \cite{zhu2020beyond, suresh2021breaking}, and layer-wise operations \cite{xu2018representation, chen2020simple, chien2020adaptive} to optimize node representation quality. 
It's worth emphasizing that our work shares similarities with that of \cite{wang2023heterophily}, as both approaches utilize proxy models to discern heterogeneity. However, our objective is specifically geared towards pruning the receptive fields that influence aggregation, granting our approach greater versatility. Additionally, our method can better aid in model storage and expedite training.

\section{Proof}\label{app:proof}

In this subsection, we present a detailed proof. The notation $\\  \mathbb{E}_{\mathbf{A} \sim SBM(p,q)}[ \cdot ]$ represents the expected pattern of the adjacency matrix. As depicted in main part, considering in subsequent layers, the HES algorithm is applied such that $p$ remains constant while heterophilic nodes are pruned, thus reducing $q$, the product of the smallest eigenvalues of $\mathbb{E}_{\mathbf{A} \sim \psi(N,p,q)} [\mathbf{G}]$ is given by:
\begin{equation}\small
    {\rm{\Pi }}_{i = 1}^L\frac{{1 - {p^{(i)}}}}{{\left( {N - 1} \right){p^{(i)}} + N{q^{(i)}} + 1}} = {\rm{\Pi }}_{i = 1}^L\left( {1 - \frac{{N\left( {{p^{(i)}} + {q^{\left( i \right)}}} \right)}}{{\left( {1 - {p^{(i)}}} \right) + N\left( {{p^{(i)}} + {q^{\left( i \right)}}} \right)}}} \right)
\end{equation}

Consider the case \( p > q \) are both functions of \( N,i \), and suppose $p=1/((N-1)*i^2),q=1/(N*i^2)$. In this scenario, as the layer depth increases, both \( p \) and \( q \) undergo decay, yet maintain the condition \( p > q \). Consequently, the product of the smallest eigenvalues can be expressed as:

\begin{equation}
    {\rm{\Psi }}\left( L \right) = {\rm{\Pi }}_{i = 1}^L\frac{{\left( {N - 1} \right){i^2} - 1}}{{\left( {N - 1} \right){i^2} + 2\left( {N - 1} \right)}}
\end{equation}


While the network goes infinitely deep, \textit{a.k.a}, $L\rightarrow \infty$, the infinite product of the smallest eigenvalues can be calculated as follows:

\begin{equation}\footnotesize
    \begin{aligned}
    & {\rm{\Psi }}\left( L \right) = {\rm{\Pi }}_{i = 1}^L\frac{{\left( {N - 1} \right){i^2} - 1}}{{\left( {N - 1} \right)({i^2}+2)}} \\
    & = { \sqrt{2} \pi \operatorname{csch}(\sqrt{2}\pi) (1 - \frac{1}{\sqrt{N-1}})_{L} (1 + \frac{1}{\sqrt{N-1}})_{L}}
    \frac{1}{\Gamma(L - i\sqrt{2} + 1)\Gamma(L + i\sqrt{2} + 1)} \\
    & \approx \sqrt{2N} \operatorname{csch}(\sqrt{2} \pi) \sin(\pi/\sqrt{N})
    \end{aligned}
\end{equation}
where $\operatorname{csch}(\cdot)$ denotes the hyperbolic cosecant function, $\Gamma(\cdot)$ denotes the Gamma function and $(a)_L = a(a+1)...(a+L-1)$ denotes the Pochhammer Symbol. When the network goes infinitely deep, the value of the infinite product asymptotically approaches a non-zero value $ \sqrt{2N} \operatorname{csch}(\sqrt{2} \pi) \sin(\pi/\sqrt{N}) $, which concludes the proof.

We posit that this phenomenon is not solely confined to binary classification contexts. Even in multi-class scenarios, a similar pattern is observed. Utilizing the HES algorithm, we can adeptly facilitate the divergence of the GNTK, as opposed to its convergence to zero, thereby enhancing the efficacy of network training.

\end{document}